%% file: elsarticle-template-num-names.tex
\documentclass[preprint,review,12pt]{elsarticle}

\usepackage{amssymb}
\usepackage{subfigure}
\usepackage{enumerate}
\usepackage{multirow}
\usepackage{adjustbox}
\usepackage{booktabs}
\usepackage{color}
\usepackage{amsmath}
\usepackage{graphicx}
\journal{Neurocomputing}

\begin{document}

\begin{frontmatter}



\title{GRSDet: Learning to Generate Local Reverse Samples for Few-shot Object Detection}

\author[label1]{Hefei Mei\corref{mycorrespondingauthor}}
\cortext[mycorrespondingauthor]{Corresponding author}\ead{hfmei@std.uestc.edu.cn}

\author[label1]{Taijin Zhao}
\author[label1]{Shiyuan Tang}
\author[label1]{Heqian Qiu}
\author[label1]{Lanxiao Wang}
\author[label1]{Minjian Zhang}
\author[label1]{Fanman Meng}
\author[label1]{Hongliang Li}
\affiliation[label1]{organization={University of Electronic Science and Technology of China},
            addressline={},
            city={Chengdu},
            postcode={611731},
            state={},
            country={China}}

% \affiliation[label2]{organization={},

\begin{abstract}

Few-shot object detection (FSOD) aims to achieve object detection only using a few novel class training data. Most of the existing methods usually adopt a transfer-learning strategy to construct the novel class distribution by transferring the base class knowledge. However, this direct way easily results in confusion between the novel class and other similar categories in the decision space. To address the problem, we propose generating local reverse samples (LRSamples) in Prototype Reference Frames to adaptively adjust the center position and boundary range of the novel class distribution to learn more discriminative novel class samples for FSOD. Firstly, we propose a Center Calibration Variance Augmentation (CCVA) module, which contains the selection rule of LRSamples, the generator of LRSamples, and augmentation on the calibrated distribution centers. Specifically, we design an intra-class feature converter (IFC) as the generator of CCVA to learn the selecting rule. By transferring the knowledge of IFC from the base training to fine-tuning, the IFC generates plentiful novel samples to calibrate the novel class distribution. Moreover, we propose a Feature Density Boundary Optimization (FDBO) module to adaptively adjust the importance of samples depending on their distance from the decision boundary. It can emphasize the importance of the high-density area of the similar class (closer decision boundary area) and reduce the weight of the low-density area of the similar class (farther decision boundary area), thus optimizing a clearer decision boundary for each category. We conduct extensive experiments to demonstrate the effectiveness of our proposed method. Our method achieves consistent improvement on the Pascal VOC and MS COCO datasets based on DeFRCN and MFDC baselines.

\end{abstract}

% % \includegraphics{grabs}

%Research highlights
% \begin{highlights}
% \item Few-shot Object Detection based on knowledge transfer easily results in confusion between the novel class and other similar categories in the decision space.
% \item When directly augmenting the sample of novel classes, it is easy to exacerbate the aliasing of feature distributions.
% \item We propose two plug-and-play models to generate and optimize the low-confused samples without introducing other pre-trained networks and extra datasets.
% \end{highlights}


\end{frontmatter}


\section{Introduction}
% \label{}

General object detection networks train on huge amounts of annotated data, which expends great cost and drags to adapt to rapidly changing application scenarios. In order to make networks draw inferences about other cases from one instance like humans, researchers focus on few-shot object detection (FSOD)~\cite{ wang2020frustratingly, yan2019meta, karlinsky2019repmet, qiao2021defrcn, leng2022sampling, du2022augmentative} to implement effectively recognition of data scarcity classes.

\begin{figure}[!t]
    \centering
    \subfigure[General variance augmentation]{
        \includegraphics[width=0.4\linewidth]{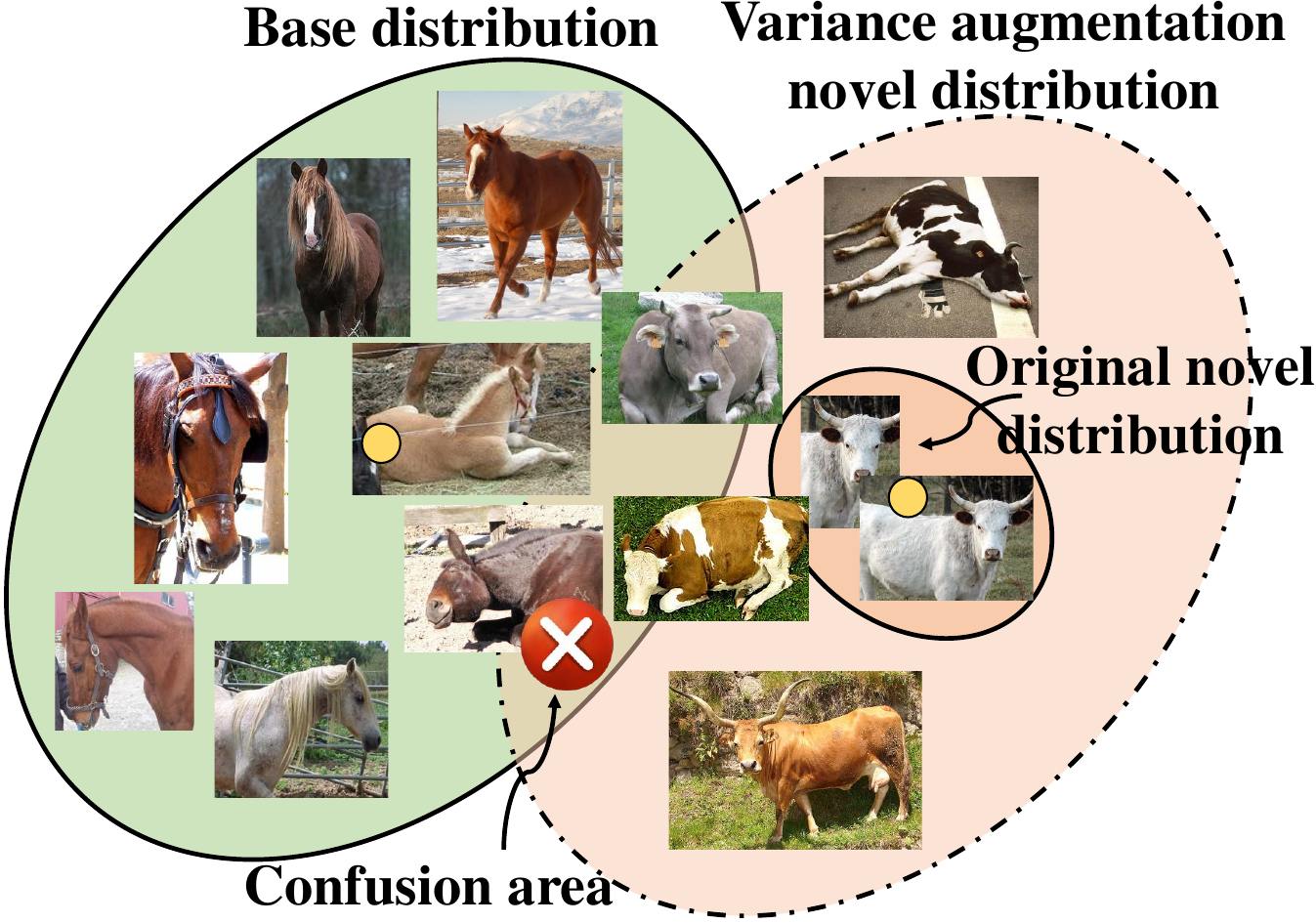}
        \label{General VA}
    }
    \subfigure[Center Calibration Variance Augmentation]{
	   \includegraphics[width=0.5\linewidth]{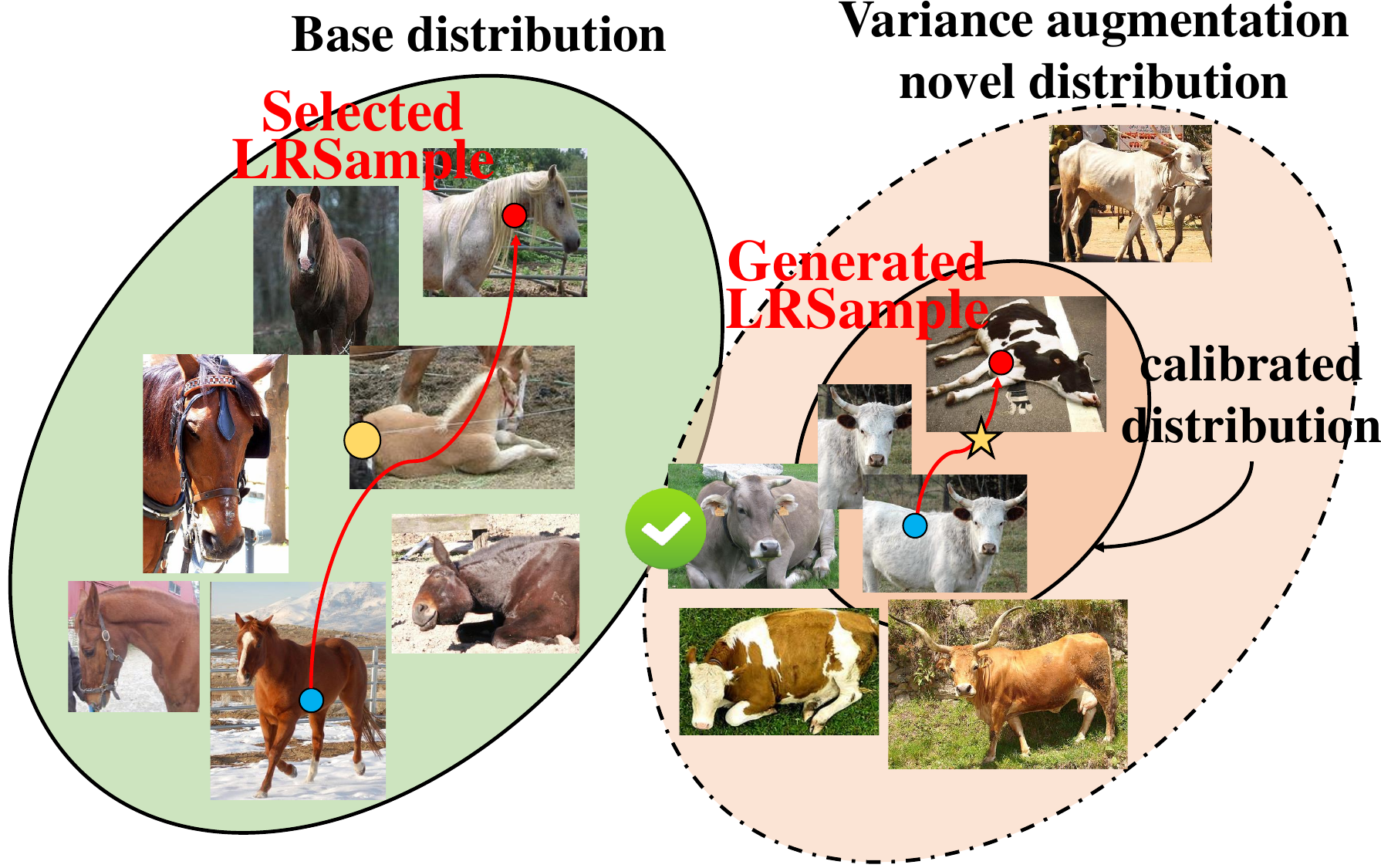}
        \label{CeCa VA}
    }
    \quad    %用 \quad 来换行
    \subfigure[Feature Density Boundary Optimization]{
    	\includegraphics[width=0.9\linewidth]{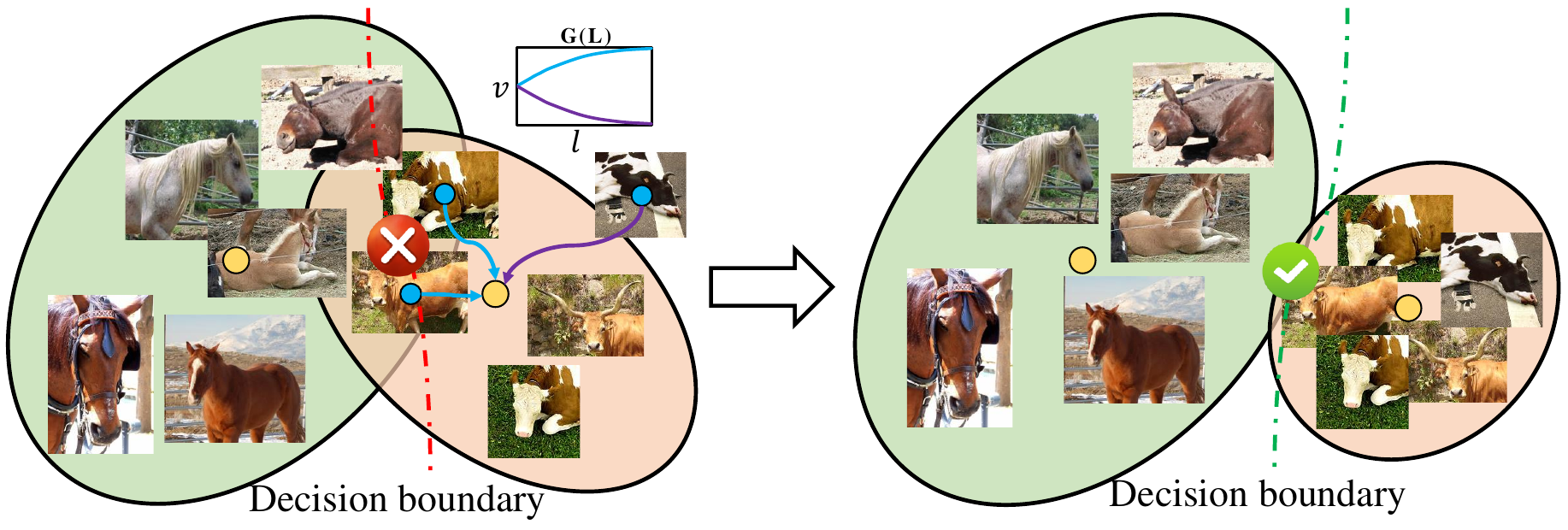}
        \label{Loss RW}
    }
    \caption{Comparison of traditional strategy and our motivation. (a) The light orange space surrounded by dashed lines is an augmented space, while the deep orange space surrounded by solid lines is the original space. The yellow dots represent the centers of each category. (b) The solid red line represents the conversion between samples achieved through an intra-class feature converter (IFC). The red pentagram is the calibrated sample center in the novel class space. (c) The dashed line represents the decision boundary between different categories, and different colors represent different importance and reweighting functions $\mathcal{G}(\cdot)$.}
    \label{Introduction}
\end{figure}

With various background information and the more complex feature extraction process, it can be difficult to establish appropriate feature distribution space in FSOD~\cite{sun2021fsce, zhang2022meta, cao2021few}. To overcome the problem, there have been many excellent works in optimizing decision space, such as EEKT~\cite{zhao2022exploring} uses ImageNet~\cite{deng2009imagenet} calibrate sample features. FADI~\cite{cao2021few} applies WordNet~\cite{miller1995wordnet} to associate and discriminate novel class features. MFDC~\cite{wu2022multi} transfers the variance of base class features to novel feature distribution in decision space.

Although the above methods have made progress in constructing novel class features, they are all based on the original novel class distribution, which is easily confused with similar base classes, causing learning difficulties in the fine-tuning stage. Concretely, the approach of utilizing feature sampling via Gaussian distribution, like ~\cite{wu2022multi} and ~\cite{zhao2022exploring}, can augment the richness of novel class features within the decision space, but it cannot modify the global shift of spatial center position induced by feature scarcity. As shown in Figure \ref{General VA}, the randomly sampled novel class features are similar to the base class features, thus leading to potential confusion between the novel class and base class features.

In this work, we partition the feature distribution into center and boundary components and design the \textbf{G}enerating Local \textbf{R}everse \textbf{S}amples as Few-shot Object \textbf{Det}ector (GRSDet), which contains Center Calibration Variance Augmentation (CCVA) module and Feature Density Boundary Optimization (FDBO) module. It can exclusively employ the network's features in optimizing the center and boundary of decision space distribution by adding the LRSamples and reweighting the loss of each sample without introducing other pre-trained networks and extra data.

Specifically, in the CCVA module, we set a LRSamples selecting criterion based on center prototype reference frames during the base training, and then we introduce an intra-class feature converter (IFC) that can learn the mapping criterion between the input samples and LRSamples. As shown in Figure \ref{CeCa VA}, through the training of the IFC, it can generate the LRSamples, which have the same category as the input samples but have significant differences. By transferring the IFC to the fine-tuning stage, CCVA can generate novel LRSamples, which jointly calibrate the novel class centers. After center calibration, CCVA makes variance augmentation on the more accurate centers to generate low confused samples for fine-tuning. We compare the distance (Normalized Euclidean) between the centers of the novel classes and the centers of the similar base classes with and without our LRSamples feature during the fine-tuning stage on Split 3 in Table \ref{tab:distance}. It can be seen that our CCVA module reduces the confusion between the center of novel classes and similar base classes.

\begin{table}
    \caption{The distance between the center of novel classes and the center of the similar base classes w/. and w/o. LRSample sets.}
    \input{tables/distance}
    \label{tab:distance}
\end{table}

In the FDBO module, we regard the recognition error samples as boundary samples and optimize them through importance weighting. We first calculate the sample densities of each edge sample in the decision space that belongs to its category and the most similar category. Then, by comparing the density of the category to which it belongs with that of the most similar categories, we divide them into two types: those close to the decision boundary (higher density of the similar category) and those far away from the decision boundary (lower density of the similar category). Figure \ref{Loss RW} illustrates the iteration under sample importance reweighting. As the sample features near the decision boundary are more likely to be confused, their importance will increase relatively. Instead, if the density of the similar category is relatively low, we will reduce its importance. As a result, the network can dynamically adjust the learning ability of each sample and more clearly define decision boundaries. By optimizing low-confused samples, our method can obtain a clearer distribution of decision space. To conclude, the contributions of our study are as follows:

\begin{enumerate}

\item[$\bullet$] We divide the distribution of the decision space into two components: the center and the boundary. Furthermore, we propose two plug-and-play models to generate LRSamples and optimize the calibrated distributions in the decision space for FSOD.

% We construct a sample mapping method based on differences and designed an intra-class feature converter to calibrate the feature centers of novel class samples in the classification decision space.

\item[$\bullet$] Our model can adaptively optimize the decision space based on the current network features, without requiring additional datasets to be introduced.

\item[$\bullet$] The effectiveness of our proposed method is demonstrated by experiments conducted on both the PASCAL VOC dataset~\cite{everingham2010pascal} and the MS-COCO dataset~\cite{lin2014microsoft} on DeFRCN and MFDC baselines.

\end{enumerate}

\section{Related Works}

\subsection{Few-shot Image Recognition} 
In the face of limited data, general image recognition strategies have become helpless. Few-shot learning aims to realize the novel class recognition network with more generalization ability under limited samples. Meta-learning based methods~\cite{jamal2019task, finn2017model, naik1992meta, bertinetto2018meta, simon2020modulating, zhu2022mgml} intend to train the task-level knowledge so that novel class can be easily adapted on the recognition network. Metric-learning based methods~\cite{sung2018learning, snell2017prototypical, vinyals2016matching, garcia2017few, zhang2020deepemd, liu2018learning, li2020adversarial, li2019distribution, oreshkin2018tadam} analyze the feature representation in the same feature space between base and novel classes, to compute the metric distance of each class. Hallucination methods~\cite{wang2018low, hariharan2017low} generate imaginary samples by adding supervised noise to obtain more generalized novel class features. Dynamic kernel methods~\cite{xu2021learning, ma2022learning} use dynamic convolution to build an adaptive feature generator so that the novel class feature distribution can be more reliable. Similar to FSL, accurate feature distribution is also crucial in FSOD.

\subsection{Few-shot Object Detection}
Based on multiple FSL strategies, a lot of work springs up in FSOD. Meta-learning based FSOD~\cite{kang2019few, fan2020few, hu2021dense, zhang2021accurate, wang2019meta, yan2019meta, han2022meta, zhang2022meta} construct meta-level knowledge based on the meta-learning paradigm used for classification and location, and design matching strategies to detect novel classes. MetaYOLO~\cite{kang2019few} uses a reweighting module to extract general features from the novel class support set, thereby exploiting the correlation of meta-features with the query set for detection. Meta R-CNN~\cite{yan2019meta} obtains meta-knowledge by extracting category attention vectors and fusing them with corresponding proposal features to achieve detection. Metric-learning based FSOD~\cite{karlinsky2019repmet, osokin2020os2d, hsieh2019one, cao2021few, li2021beyond} seek suitable mapping space to enhance the separability of novel class features following metric-learning paradigm. CME~\cite{li2021beyond} changes the classification space margin by disturbing novel class features to avoid confusion between classes. Transfer-learning based FSOD~\cite{chen2018lstd, wang2020frustratingly, qiao2021defrcn, sun2021fsce, zhang2021hallucination, li2021few, wu2021universal, zhu2021semantic, zhao2022exploring, wu2022multi} receive more attention recently due to its simple network structure and good performance. TFA~\cite{wang2020frustratingly} first proposes an effective transfer-learning framework via a freezing feature extractor when fine-tuning. DeFRCN~\cite{qiao2021defrcn} decouples the learning rate of RPN and R-CNN. FSCE~\cite{sun2021fsce} reduces the confusion of novel classes in the feature space by incorporating contrastive learning. HallucFsDet~\cite{zhang2021hallucination} enriches novel class features by adding anti-noise vectors to the adversarial network. MFDC~\cite{wu2022multi} establishes the semantic relationship between the base class and novel class features in a distillation manner explicitly.

% TODO：对比学习、对抗学习

\subsection{Distribution Optimization of Decision Space}

The distribution of features in the decision space has a significant impact on the results of the network. Comparative learning~\cite{grill2020bootstrap, he2020momentum} aims to learn a mapping of feature space that brings samples of the same category closer while pushing samples of different categories farther apart. By doing so, the method can construct a clearer distribution of decision spaces. Generative Adversarial Network~\cite{goodfellow2020generative} produces realistic samples, which can make the same-class samples more clustered. FSOD methods~\cite{sun2021fsce, wang2018low, zhang2021hallucination} achieve good few-shot performance using the above ideas. Hard example mining~\cite{shrivastava2016training} believes that emphasizing misidentified samples during training can improve the network's ability to fit such samples. Curriculum learning methods~\cite{wang2021survey, kumar2010self, jiang2015self} enable the network to achieve better generalization by gradually learning from simple samples to hard samples during the training process. FSOD method~\cite{kang2019few} uses feature reweighting to establish meta-knowledge. Motivated by the above arguments, we propose distribution optimization for both center and edge samples. Compared with existing methods, our method can generate LRSamples directionally, which enhances sample richness and calibrates the distribution of novel classes through a simple structure.

\section{Proposed Method}
\label{method}

In this section, we design an FSOD model that calibrates and optimizes the decision space using only a single dataset and a simple but effective structure. Firstly, we present the preliminary of FSOD in Section \ref{section_Prelimnary}. Then we propose the CCVA module in Section \ref{section_calibration}, which can transfer the LRSamples generating information from the base to the novel classes to calibrate the novel class distribution. The CCVA will also take variance augmentation to sample more novel features for the improvement of novel sample richness. Furthermore, we introduce the FDBO module in Section \ref{section_reweight}, which can reweight the importance of edge samples to optimize the boundary of novel distribution in the decision space. Finally, we provide the overall training loss of our method.

\begin{figure*}[ht]
\centerline{\includegraphics[width=1\linewidth]{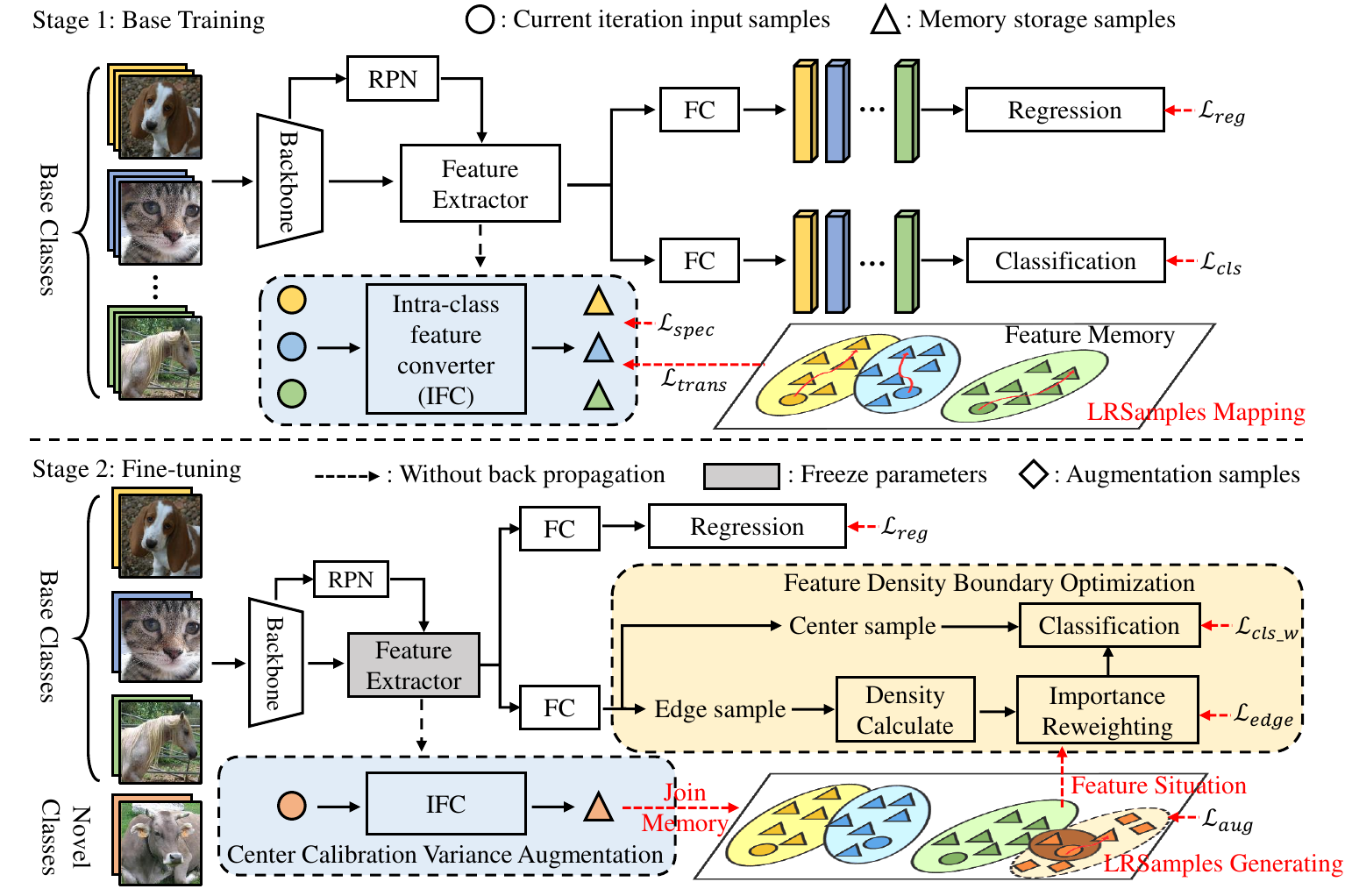}}
	\caption{Overview of our proposed GRSDet. The Center Calibration Variance Augmentation (CCVA) module implements LRSamples generated through learning on base training and transferring on fine-tuning, then uses similar base class variance to augment the calibrated novel feature. The Feature Density Boundary Optimization (FDBO) module judges the distance from the decision boundary by calculating the sample density to design the weight of importance. More details can be referred to in Section \ref{method}.}
	\label{overview}
\end{figure*}

\subsection{Preliminaries}
\label{section_Prelimnary}

\textbf{Problem Definition.} \enspace The FSOD setting in our paper follows the previous works~\cite{wang2020frustratingly, kang2019few}. Specifically, we split the classes into two sets: base classes $C_{base}$ and novel classes $C_{novel}$, where the novel sets only contain $\emph{K}$ instances (usually $\emph{K} \leq 30$) per category and $C_{base}\cap C_{novel}=\varnothing$. Following the paradigm of transfer-learning, we first train our network on $C_{base}$ and then fine-tune on $C_{novel}$, finally, we test the overall network on $C_{base}\cup C_{novel}$. Based on existing data stream~\cite{wang2020frustratingly, wu2022multi}, we give fine-tuning results on both balanced training set $D = D_{base}\cup D_{novel}$.

As for the fine-tuning stage, different from the existing transfer-learning FSOD~\cite{qiao2021defrcn, wu2022multi, wang2020frustratingly, sun2021fsce}, which backpropagate average loss of each sample to optimize the network $\mathcal{F}(I_i,y_i,\theta|\theta_{base})$, our method reweights the sample importance based on the distribution status of the feature space. Particularly, we define an importance weight function $\mathcal{G}(\cdot)$ and introduce a suitable training strategy, which can be expressed as:

\begin{equation}
\begin{aligned}
    \underset{\theta}{\min}\sum_i{v_i\mathcal{L}\left( I_i,y_i,\theta |\theta _{base} \right)} + \mathcal{L}\left( v_i \right)\\
    s.t.\ v_i=\mathcal{G}\left( l_i\left( I_i,y_i,\theta |\theta _{base} \right) \right)
\end{aligned}
\label{overformula}
\end{equation}
where $I_i$ is the inputs images, $y_i$ is the corresponding labels and $\mathcal{L}$ is the overall loss during fine-tuning. The network optimizes the parameters $\theta$ and weights based on the parameters of base training $\theta_{base}$.

\noindent\textbf{Framework of Few-shot Detector.} \enspace Our FSOD framework is illustrated in Figure \ref{overview}, which adopts the two-stage fine-tuning paradigm and widely used object detector Faster R-CNN~\cite{ren2015faster}. In the base training stage, we train the network on the base classes with $\mathcal{L}_{reg}$ and $\mathcal{L}_{cls}$. By searching for differential mapping samples for each input sample on the feature memory bank proposed in ~\cite{wu2022multi}, we supervise the intra-class feature converter in the CCVA module with $\mathcal{L}_{spec}$ and $\mathcal{L}_{trans}$. In the fine-tuning stage, we transfer the knowledge of CCVA to generate novel class LRSamples, to calibrate the novel feature distribution. Moreover, we select edge samples by recognition accuracy and calculate the sample density of self class and similar class to divide their spatial position situation. Then the importance reweighting is applied with the classifier, weighting the edge samples loss with balance supervision $\mathcal{L}_{edge}$ in the FDBO module.

\subsection{Center Calibration Variance Augmentation}
\label{section_calibration}

In this section, we design a selection criterion for LRSamples in the prototype reference frame of each class, which can transform the input sample features into other intra-class sample features with relatively low similarity. The CCVA trains the intra-class feature converter (IFC) under the supervision of criterion during the base training and then transfers the knowledge to generate novel LRSamples to calibrate the distribution in decision space during the fine-tuning stage. Afterward, CCVA performs variance augmentation on the calibrated center, which contains less confusion in network decision-making during fine-tuning.

\begin{figure}
	\centerline{\includegraphics[width=0.75\linewidth]{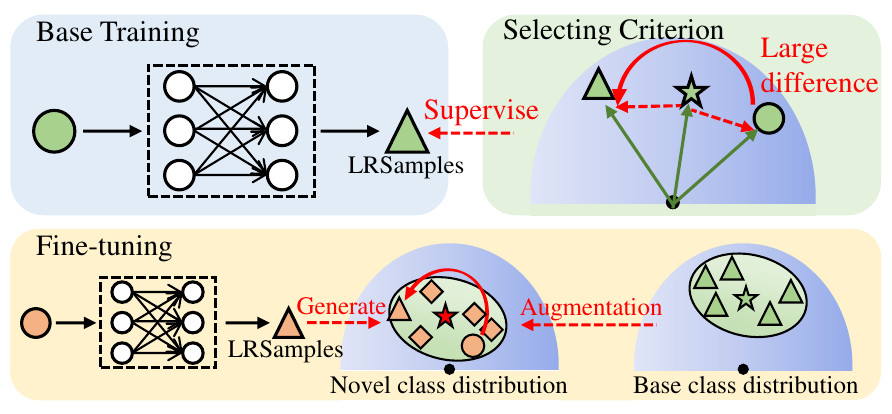}}
	\caption{Structure of Center Calibration Variance Augmentation module. It contains the Selection Criterion of LRSamples, IFC on base training and fine-tuning stages, and variance augmentation three processes. The type of sample shape is the same as Figure \ref{overview}.}
	\label{CCVA}
\end{figure}

\noindent\textbf{Selecting Criterion of LRSamples.} \enspace In the base training stage, we use the memory bank to get a rich feature decision space. Different from ~\cite{wu2022multi}, which obtains prototypes based on the entire feature information of each class, we delve into a sample-level aspect of the usage of the memory bank. Concretely, we establish differential mapping relationships, which is shown in the mapping process of Figure \ref{CCVA}. In order to unify the measurement of difference vectors, we place all vectors on a normalized hypersphere. Then we can calculate the difference between each feature and the corresponding prototype as follows so that the sample relationships will be converted to the prototype reference frame(Red dashed line in Figure \ref{CCVA}):

\begin{equation}
\begin{aligned}
    D_f\left( x_{in} \right) =\hat{x}_{in}-\hat{P}_{Cin}
\end{aligned}
\label{diff_in}
\end{equation}

\begin{equation}
\begin{aligned}
    D_f\left( x_{bank}^{i} \right) =\hat{x}_{bank}^{i}-\hat{P}_{Cin},\ i\in \left\{ 1,2,\cdots ,N_{Cin} \right\} 
\end{aligned}
\label{diff_bank}
\end{equation}
where $\hat{x}_{in}$ is the normalized features of input images, $\hat{P}_{Cin}$ is the normalized prototype of class $Cin$ and $\hat{x}_{bank}^{i}$ is each normalized features in the memory bank of class $Cin$. The length of it is $N_{Cin}$.

After getting the difference vectors of input features and memory bank features, we compute the similarity between those two kinds of vectors. Afterward, we obtain the order of differences through sorting:

\begin{equation}
\begin{aligned}
 \mathcal{A} = \underset{i\in \left\{ 1,\cdots N_{Cin} \right\}}{\text{argsort}}\frac{D_f\left( x_{in} \right) \cdot D_f\left( x_{bank}^{i} \right)}{\lVert D_f\left( x_{in} \right) \rVert \lVert D_f\left( x_{bank}^{i} \right) \rVert}
\end{aligned}
\label{diff}
\end{equation}

To prevent only differential mapping from capturing edge hard samples, we increase the similarity evaluation with the prototypes, and the formula is as follows:

\begin{equation}
\begin{aligned}
    D_c\left( x_{in} \right) =\frac{\hat{x}_{in}\cdot \hat{P}_{Cin}}{\lVert \hat{x}_{in} \rVert \lVert \hat{x}_{in} \rVert}
\end{aligned}
\label{diss_in}
\end{equation}

\begin{equation}
\begin{aligned}
    D_c\left( x_{bank}^{i} \right) =\frac{\hat{x}_{bank}^{i}\cdot \hat{P}_{Cin}}{\lVert \hat{x}_{bank}^{i} \rVert \lVert \hat{x}_{in} \rVert},\ i\in \left\{ 1,2,\cdots N_{Cin} \right\} 
\end{aligned}
\label{diss_bank}
\end{equation}

Similar to the assessment of differences, we rank the similarities:

\begin{equation}
\begin{aligned}
 \mathcal{B} = \underset{i\in \left\{ 1,\cdots ,N_{Cin} \right\}}{\text{argsort}}\left| D_c\left( x_{in} \right) -D_c\left( x_{bank}^{i} \right) \right|
\end{aligned}
\label{diss}
\end{equation}

Finally, we use softmax to get the probability of fit to our requirements. By synthesizing the results of formula \ref{diff} and formula \ref{diss}, we search for the final mapping feature location:
\begin{equation}
\begin{aligned}
l = \underset{i\in \left\{ 1,\cdots ,N_{Cin} \right\}}{\text{arg}\min}p_{\mathcal{A}} + p_{\mathcal{B}}
\end{aligned}
\end{equation}
where $\mathcal{A}$ and $\mathcal{B}$ are the different criteria for selecting LRSamples, $p$ indicates the probability of fit to our requirements, which is getting from two sort processes.

\noindent\textbf{Base Training Stage.} \enspace We design an intra-class feature converter (IFC) to generate LRSample features under the Selecting Criterion, which is composed of a Multi-Layer Perceptron with a single hidden layer. As shown in the base training stage of the Figure \ref{CCVA}, we use IFC to obtain transfer features $x_{trans}^i$ and use the target features $x_{target}^i$ obtained from the mapping relationship for supervision. The correctness supervision loss of transfer features is as follows:

\begin{equation}
\begin{aligned}
 \mathcal{L}_{trans}=\frac{1}{n}\sum_i{\frac{x_{trans}^{i}\cdot x_{target}^{i}}{\lVert x_{trans}^{i} \rVert \lVert x_{target}^{i} \rVert}}
\end{aligned}
\label{loss_trans}
\end{equation}

Meanwhile, by adding cross-entropy loss, we ensure that the generated features are specific-class:

\begin{equation}
\begin{aligned}
 \mathcal{L}_{spec}=\mathcal{L}_{CE}\left( \mathcal{F}_{cls}\left( x_{trans}^{i} \right) ,Cin \right) 
\end{aligned}
\label{loss_spec}
\end{equation}
where $\mathcal{F}_{cls}$ is the classifier of the network.

\noindent\textbf{Fine-tuning Stage.} \enspace We use the IFC to generate novel class LRSamples. Then we update the class feature center by adding these samples to the memory bank. As for the variance augmentation, we follow the method in MFDC~\cite{wu2022multi} to transfer the Gaussian distribution variance and sample randomly:

\begin{equation}
\begin{aligned}
\mathbf{N}_{novel}=\left\{ x|x\sim \mathcal{N}\left( \mu _{novel},\frac{1}{k}\sum_i{\sigma _{base}^{i}} \right) \right\} 
\end{aligned}
\label{augmentation}
\end{equation}
where $\mathbf{N}_{novel}$ is the distribution of a novel class, and $\mathcal{N}$ is the Gaussian distribution. Finally, we use the sample center in the calibrated distribution $\mathbf{N}_{novel}$ and follow MFDC~\cite{wu2022multi} for variance augmentation, setting loss for the augmented samples as follows:

\begin{equation}
\begin{aligned}
\mathcal{L}_{aug}=\mathcal{L}_{CE}\left( \mathcal{F}_{cls}\left( x_{aug}^{i} \right) ,Cin \right) 
\end{aligned}
\end{equation}
where the $x_{aug}^{i}$ is the augmented sample obtained by using similar base class variances for migration according to MFDC~\cite{wu2022multi}.

\subsection{Feature Density Boundary Optimization}
\label{section_reweight}

\begin{figure}
	\centerline{\includegraphics[width=0.75\linewidth]{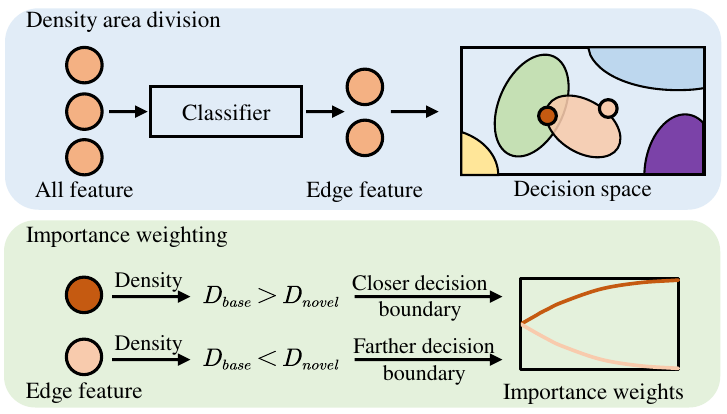}}
	\caption{Structure of Feature Density Boundary Optimization module. Dark orange and light orange features belong to the same category and have different importance depending on the density.}
	\label{FDBO}
\end{figure}

We divide the feature distribution into centers and boundaries. In this section, we optimize the distribution boundaries in the decision space as shown in Figure \ref{FDBO}. Firstly, we use the normal classifier to output error samples and use them as decision edge samples for density calculation. Then, by comparing the sample density of the input category and similar category, we classify its importance. Finally, the importance weighting function $\mathcal{G}(\cdot)$ is used to weight the classification loss.

\noindent\textbf{Density area division.} \enspace We differentiate regions by comparing the sample density of the input sample category and the most similar category, where the similar category is calculated as:

\begin{equation}
\begin{aligned}
C_{sim}=\underset{i\in \mathbb{Z}_C,i\ne Cin}{\text{arg}\min}\lVert x_{in}-P_i \rVert 
\end{aligned}
\end{equation}
where $\mathbb{Z}_C$ is all categories during fine-tuning. Similar to the density peak clustering method, we use distance to calculate the number of samples in a certain region as its regional density. Due to inconsistent distribution among different categories, we use parameters $\eta$ To adjust the distance threshold for similar categories. The density calculation formula for the two categories is as follows:

\begin{equation}
\begin{aligned}
d_{sim}=\frac{\left| \mathbf{N}^d\left( x_{C_{sim}} \right) \right|}{\left| \mathbf{N}\left( x_{C_{sim}} \right) \right|}, \mathbf{N}^d\left( x_{C_{sim}} \right) =\left\{ x|dist\left( x,P \right) <\eta d_{thred} \right\} 
\end{aligned}
\end{equation}
where $d_{thred}$ is the distance in input category, which make the $d_{in} = 0.3$, and we choose $\eta=1.5$ in this paper.

By comparing the density values, we divide the edge samples into two sets to reflect the distance from the decision boundary. Among them, areas with a high density of similar categories are relatively important near the Decision boundary, and vice versa. The two sets can be expressed as follows:

\begin{equation}
\begin{aligned}
\mathcal{I}=\left\{ x|d_{sim}\left( x \right) >d_{in}\left( x \right) \right\} \rightarrow High\ importance
\end{aligned}
\label{area_important}
\end{equation}

\begin{equation}
\begin{aligned}
\mathcal{D}=\left\{ x|d_{sim}\left( x \right) <d_{in}\left( x \right) \right\} \rightarrow Low\ importance
\end{aligned}
\label{area_lowimportant}
\end{equation}

\noindent\textbf{Importance weighting of FDBO.} \enspace We use the loss result obtained from the normal classifier as the independent variable of importance weight and design multiple weighting functions $\mathcal{G}(\cdot)$ for importance calculation. The expression of the weighting function is shown in Table \ref{tab:Gx}. We design a loss function $\mathcal{K}$ for the weight vector $v$ to make the weight vector stable. Since most central samples have a weight value of 1 and $\mathcal{I}\cap \mathcal{D}={1}$, the loss will enable better bundling of edge samples, which is expressed as:

\begin{equation}
\begin{aligned}
\mathcal{L}_{edge}=\frac{1}{n}\sum_i{\frac{\lVert v_i-mean\left( v \right) \rVert}{\lVert v_i \rVert}}
\end{aligned}
\end{equation}

\begin{table}
    \caption{Different function designs for $\mathcal{G}(\cdot)$.}
    \input{tables/GX}
    \label{tab:Gx}

\end{table}

Finally, we calculate the weighted cross entropy loss by combining the importance vector and the loss output by the normal classifier as the classification loss weight:

\begin{equation}
\begin{aligned}
\mathcal{L}_{cls\_w}=\frac{1}{n}\sum_i{v_i\cdot \mathcal{L}_{CE}\left( \mathcal{F}_{cls}\left( x^i \right) ,Cin \right)}
\end{aligned}
\end{equation}
where the $\mathcal{L}_{CE}$ means the cross entropy loss, $v_i$ is calculated by the constraint conditions given in formula \ref{overformula}.

\subsection{Overall Training Loss}

In the base training stage, We add supervision to the CCVA module on top of the Faster Rcnn basic detector. The base class training loss can be expressed as follows:

\begin{equation}
\begin{aligned}
\mathcal{L}_{base}=\mathcal{L}_{cls}+\mathcal{L}_{reg}+\lambda_{1} \cdot \mathcal{L}_{trans}+\lambda_{2} \cdot \mathcal{L}_{spec}
\end{aligned}
\end{equation}
where the $\lambda_{1}=0.05$ and $\lambda_{2}=0.4$ in our method. Then we use CCVA and FDBO modules to jointly optimize the feature distribution in the decision space during the fine-tuning stage, the total loss can be represented as:

\begin{equation}
\begin{aligned}
\mathcal{L}_{novel}=\mathcal{L}_{cls}+\mathcal{L}_{reg}+\mathcal{L}_{cls\_w}+\lambda_{3} \cdot \mathcal{L}_{edge}+\lambda_{4} \cdot \mathcal{L}_{aug}  
\end{aligned}
\end{equation}
where the $\lambda_{3}=0.3$ and $\lambda_{4}=0.1$ in our method.

\section{Experiments}

\subsection{Experimental Setup}

\noindent\textbf{Datasets.} \enspace We evaluate our method on both Pascal VOC~\cite{everingham2010pascal} and MS COCO~\cite{lin2014microsoft} benchmark datasets. For Pascal VOC, we split it into 5 novel classes and 15 base classes following the previous work~\cite{kang2019few, wang2020frustratingly, qiao2021defrcn}. In the fine-tuning stage, we choose the same number of $k$ ($k=1,2,3,5,10$ in Pascal VOC) shot from rich base class images following ~\cite{wu2022multi}. For MS COCO, the 80 categories are split into 20 novel classes and 60 base classes, which allocate $k=1,2,3,5,10,30$ samples respectively in the fine-tuning stage.

\begin{table}[t]
    \caption{Comparison results on PASCAL VOC. We evaluate performance ($nAP_{50}$) on three splits.}
    \input{tables/pascal_voc}

    \label{tab:pascal_voc}
\end{table}

\noindent\textbf{Implementation Details} \enspace Our GRSDet method is evaluated upon three baselines: Defrcn~\cite{qiao2021defrcn} and TFA~\cite{wang2020frustratingly}, the general transfer learning-based FSOD model, and MFDC~\cite{wu2022multi}, a state-of-the-art FSOD method. We use ResNet-101~\cite{he2016deep} pretrained on ImageNet~\cite{deng2009imagenet} as our backbone and train our detector with 16 images per batch (8 GPUs, 2 images per GPU). The learning rate is set at 0.01 at base training and 0.005 at the fine-tuning stage. Consistent with MFDC~\cite{wu2022multi}, our method uses two stream settings for fine-tuning the rich base class data. Each of our modules works after 200 iterations on Pascal VOC and 1000 iterations on MS COCO during fine-tuning, for the storage of decision space features in the memory bank. To capture more rich features for differential mapping~\cite{wu2022multi}, we set the length of the memory bank to 4096.

\subsection{Comparison Results}

\begin{table}[t]
    \caption{Results on MS COCO. We evaluate performance ($nAP$ and $nAP_{75}$) under six shot settings.}
\input{tables/ms_coco}
    \label{tab:mscoco}
\end{table}

\noindent\textbf{Results on Pascal VOC.} \enspace Table \ref{tab:pascal_voc} shows the detection results of our method on Pascal VOC~\cite{everingham2010pascal}. Our method achieves a consistent performance increase compared to the two baselines. GRSDet-DeFRCN can outperform the MFDC method with DeFRCN as the baseline in terms of overall performance average. Specifically, compared with MFDC, GRSDet-DeFRCN achieves 10.3\% (vs. 6.9\%) performance increase for 1-shot, 6.9\% (vs. 3\%) for 2-shot, 2.9\% (vs. 1.9\%) for 10 shot on Novel Set 3. As for GRSDet-MFDC, it can achieve a 4.1\% increase for 1-shot, 2.6\% for 2-shot, and 2.5\% for 10-shot on Novel Set 3. Moreover, GRSDet-DeFRCN surpasses the baseline by 6.0\% and 7.1\% on average on Novel Set 1 and Novel Set 2.

\noindent\textbf{Results on MS COCO.} \enspace Table \ref{tab:mscoco} presents the experimental results on MS COCO~\cite{lin2014microsoft} benchmark. Due to the generation of rich LRSamples, our method can converge faster to higher performance. We reduce the number of fine-tuning iterations by 400 under the two stream setting of MFDC~\cite{wu2022multi}. In low (1-5) shot situations, our method can effectively improve performance on two baselines, where GRSDet can surpass DeFRCN by 8.6\% and surpass MFDC by 4.3\%. In high (10, 30) shot situations, due to the rich features of the novel class, they can construct a better distribution. Our method can maintain the original performance. On the whole, our method still has the effect of optimizing feature space distribution on the MS COCO dataset.

\subsection{Ablation Study}

\begin{table}[t]
    \caption{Effectiveness of each functional component.}
    \input{tables/ablation}
    \label{tab:ablation}
\end{table}

\noindent\textbf{Effectiveness of each functional component.} \enspace We conduct the ablations of each proposed module in GRSDet for 1, 2, and 3 shot scenarios on Novel Set 3. In the 1shot setting, the scarcity of samples results in a severe drift in the distribution center. In this case, CCVA exhibits high performance by calibrating the center and augmenting samples. As the number of images increases, sample features become relatively richer at 3shot, but confusion between the novel and base classes persists in the decision boundary. By further optimizing the boundary, FDBO achieves better performance benefits for high shots where $\mathcal{I}$ and $\mathcal{D}$ are different reweighting strategies in Section \ref{section_reweight}. Table \ref{tab:ablation} shows that combining the two modules yields the best performance.

\begin{figure}[t]
    \centerline{\includegraphics[width=0.82\linewidth]{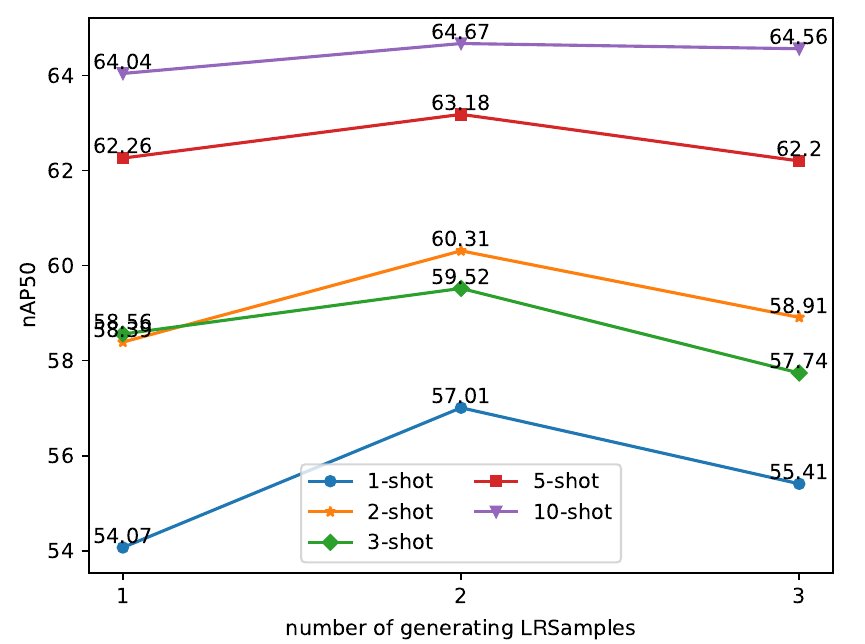}}
    \caption{Ablation of the number of generating LRSamples.}
    \label{Number trans}
\end{figure}

\noindent\textbf{Number of generating LRSamples in CCVA.} \enspace In the CCVA module, IFC can use a cascaded structure to continuously generate LRSamples and jointly calibrate the feature centers. We show the performance effects of transferring different sample sizes on 1, 2, and 3 shots under Novel Set 3 conditions in Figure \ref{Number trans}. When the number of generating LRSamples increases, due to the sample richness, better calibration of the sample center can be performed, so the effect of transferring two samples is better than that of one sample. However, when there are too many LRSamples, the noise in the LRSamples may affect the original distribution, resulting in a decrease in the effectiveness of transferring three samples. In the end, we choose to generate two LRSamples in our method.

\noindent\textbf{Effectiveness of Center Calibration in CCVA.} \enspace We perform ablation on the role of Center Calibration by LRSamples in the CCVA module on Novel Set 3. In Table \ref{tab:cceffect}, it can be seen that sample augmentation at the calibrated center can improve the performance of the network in both low and high-shot situations compared to direct augmentation. In addition, in high-shot situations, Center Calibration is more effective in calibrating the sample center, as it can improve the performance of 10-shot by $+1.1$ AP.

\begin{table}
    \caption{Effectiveness of our Center Calibration module. "w/o." indicates Variance Augmentation without Center Calibration. "w/." indicates our CCVA module.}
    \input{tables/CA}
    \label{tab:cceffect}
\end{table}

\begin{figure}	\centerline{\includegraphics[width=0.85\linewidth]{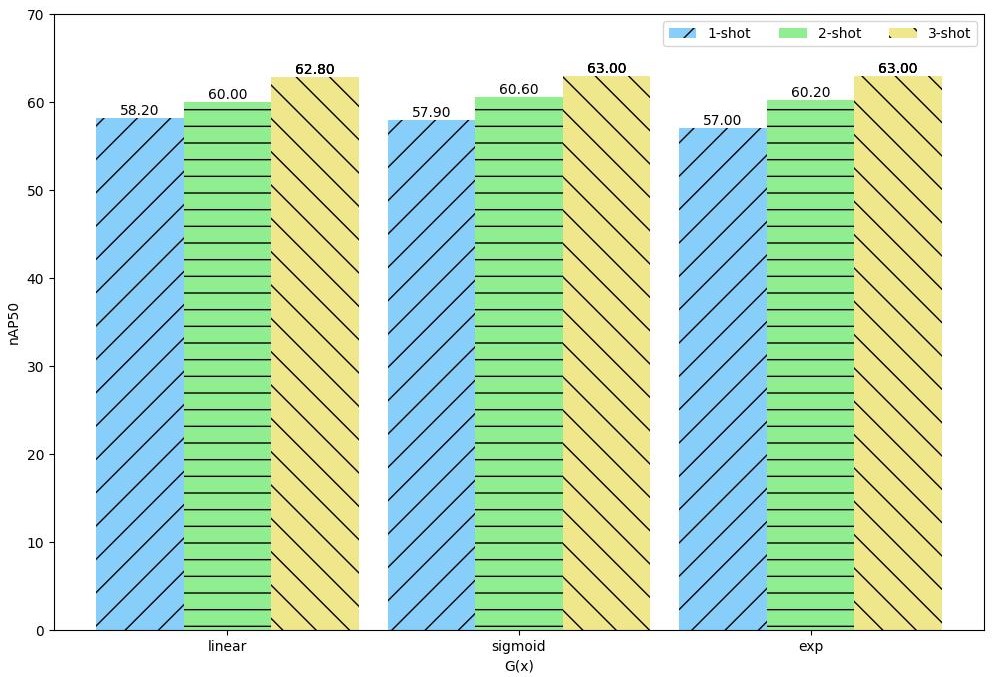}}
	\caption{Ablation of different reweighting function $\mathcal{G}(\cdot)$ in FDBO.}
	\label{Diff:function}
\end{figure}

\noindent\textbf{Different reweighting function in FDBO.} \enspace We explore the performance benefits of using the various functions listed in Table \ref{tab:Gx} for importance weighting, and Figure \ref{Diff:function} shows the performance of each function under 1, 3, and 5 shots of Novel Set 3. Each function can improve performance, indicating the generality of our FDBO module. The Linear function is more effective at low shots, but its performance is lower at high shots. The exp function performs poorly at low-shot, while the sigmoid function performs better overall. Finally, we choose the sigmoid function for the FDBO module in our method.

\begin{figure}
    \centering
    \subfigure[2-shot]{
    \includegraphics[scale=0.35]{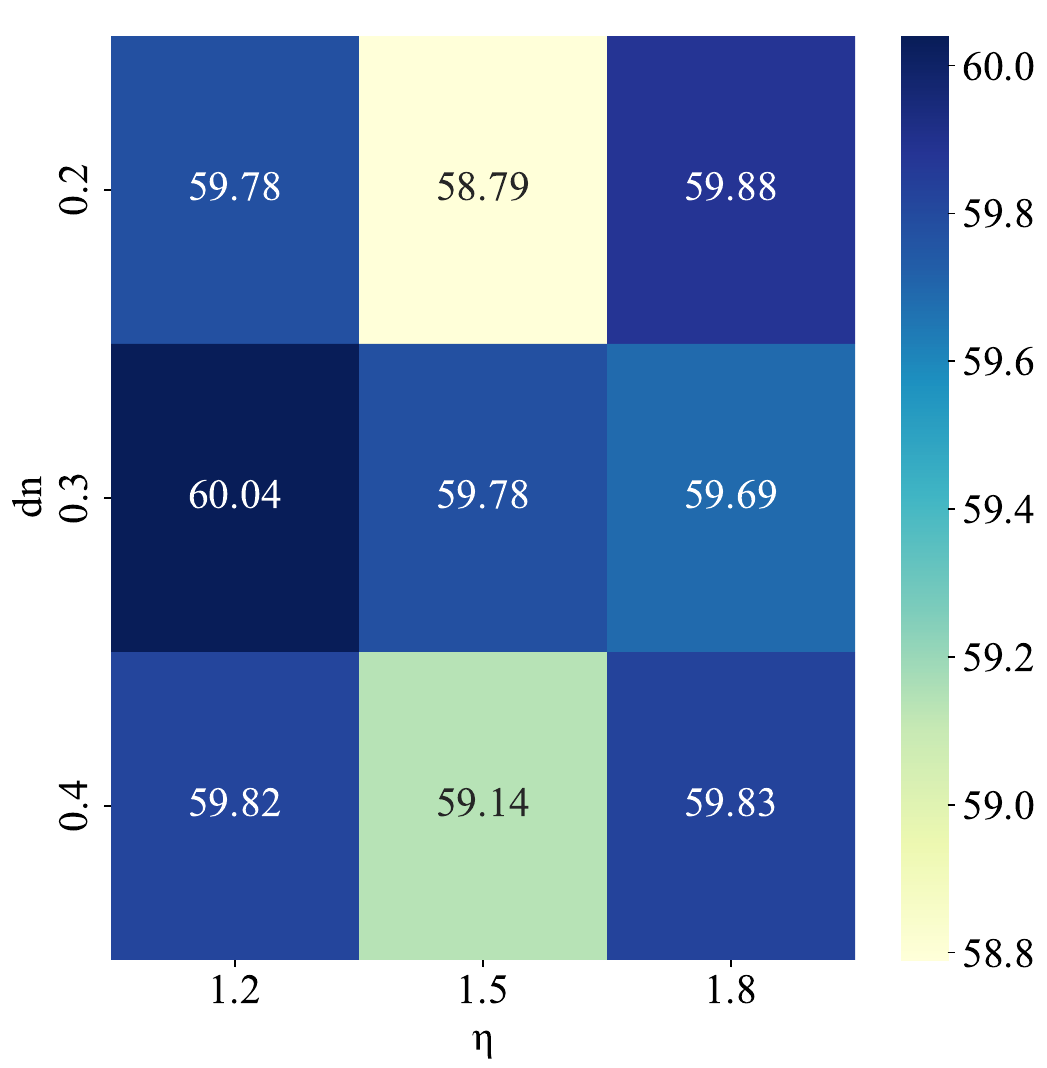}}
    \subfigure[5-shot]{
    \includegraphics[scale=0.35]{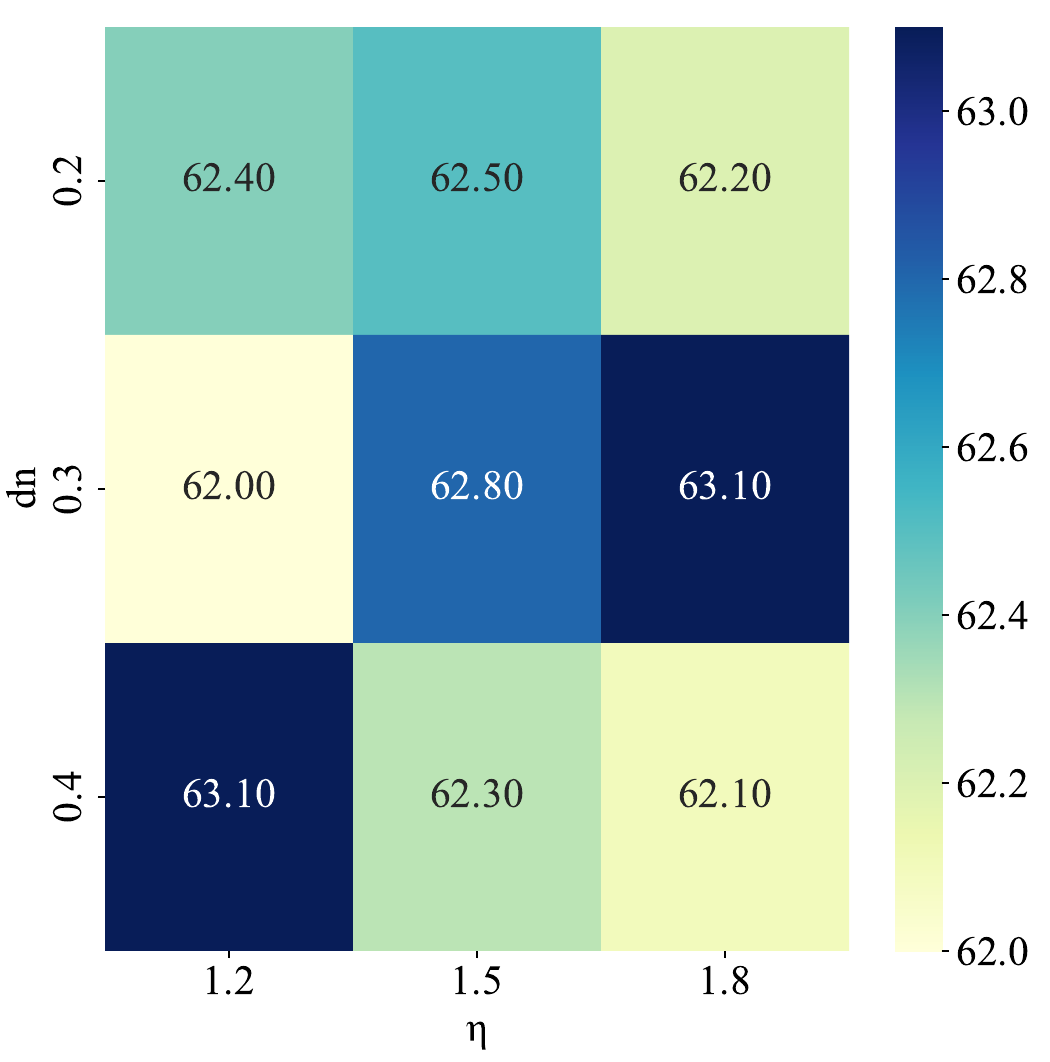}}
    \caption{Ablation of density function parameters in FDBO.}
    \label{density}
\end{figure}

\noindent\textbf{Ablation of density function parameters in FDBO.} \enspace We perform ablation on the parameters of the sample density calculation process in FDBO, and the results on Novel Set 3 are shown in Figure \ref{density}. For various parameter selections, our method shows performance improvement compared with the baseline, demonstrating its versatility. In the case of the low-shot, the effect is generally better when the novel class density $d_{in}$ is 0.3. However, in high-shot situations, results with $\eta>1.5$ are relatively good. Finally, we choose $d_{in}=0.3$ and $\eta=1.5$ in our method.

\begin{table}[t]
    \caption{Base class performance after fine-tuning.}
    \input{tables/base}
    \label{tab:base}
\end{table}

\noindent\textbf{Base class performance after fine-tuning.} \enspace We compare the base class performance with MFDC after fine-tuning, which both fine-tune on rich base class images. The results are shown in Table \ref{tab:base}, our method has better base class performance in most shots, indicating that after calibrating the novel class distribution center and optimizing its boundary, the distinctness of the base classes will also be improved.

\begin{table}[!t]
    \caption{Different loss weights in the base training stage.}
    \input{tables/loss_base}
    \label{tab:loss_b}
\end{table}

\begin{table}[!t]
    \caption{Different loss weights in the fine-tuning stage.}
    \input{tables/loss_finetune}
    \label{tab:loss_f}
\end{table}

\noindent\textbf{Ablation of loss weights.} \enspace Table \ref{tab:loss_b} shows the performance of different base training loss weights on Novel Split 3. We can obtain the best fine-tuning results when the $\lambda_{1}=0.05$ and $\lambda_{2}=0.4$, so we set the above weights for all our experiments.

We ablate the loss weights in the fine-tuning stage based on the DeFRCN baseline on Novel Split 3. From the table \ref{tab:loss_f}, it can be seen that high or low loss weights will lead to a decrease in network performance. Finally, we choose $\lambda_{3}=0.3$ and $\lambda_{4}=0.1$ for the best overall performance.

\begin{table}[!t]
    \caption{Effect of different IFC structures.}
    \input{tables/ifc}
    \label{tab:ifc}
\end{table}

\begin{table*}[!t]
    \caption{Performance of our method with different classical baselines.}
    \input{tables/baseline}
    \label{tab:baseline}
\end{table*}

\noindent\textbf{Effect of different IFC structures.} \enspace In the Center Calibration Variance Augmentation (CCVA) module, we use different Multi-Layer Perceptron structures as IFC for feature transformation. We ablate the channel dimension and activation function of the hidden layer based on the MFDC baseline on Novel Split 1. In our setting, "Low channel", "High channel", and "Equal channel" represent that the number of the hidden layer channels is half, twice, and equal to the input channel, respectively. Moreover, "LeakyReLu" denotes that we use LeakyReLu as the activation function of the hidden layer. Table \ref{tab:ifc} shows that reducing the hidden layer channel and using LeakyReLu will both reduce network performance. We finally chose "Equal channel", which performs well on both low-shot and high-shot.

\begin{figure}[!t]
    \centering
    \subfigure[DeFRCN]{
    \includegraphics[scale=0.41]{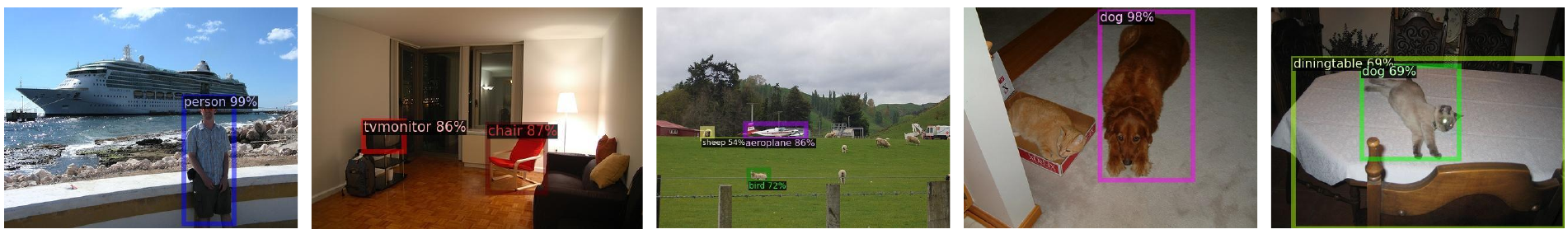}}
    
    \subfigure[GRSDet-DeFRCN]{
    \includegraphics[scale=0.41]{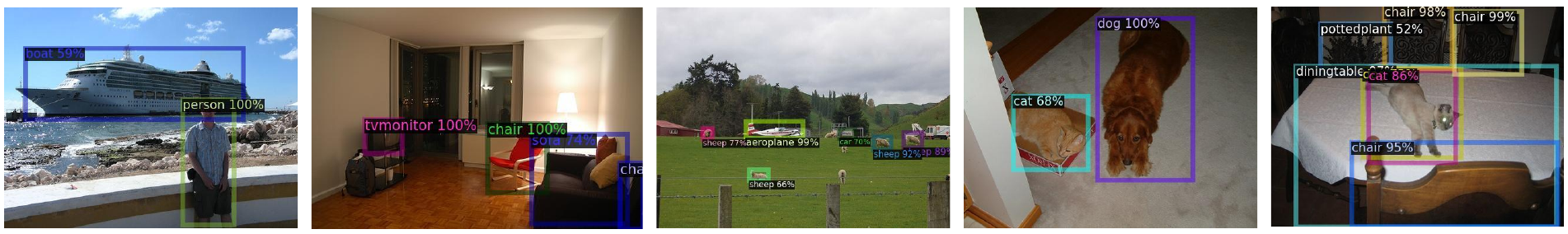}}
    
    \subfigure[MFDC]{
    \includegraphics[scale=0.41]{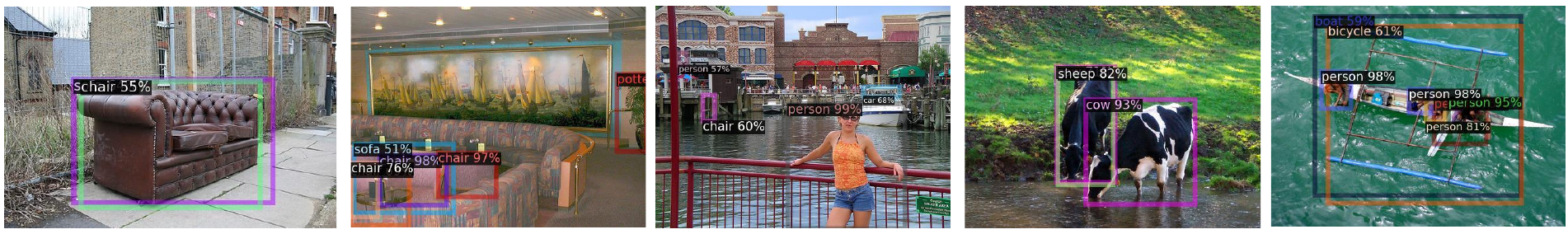}}
    
    \subfigure[GRSDet-MFDC]{
    \includegraphics[scale=0.41]{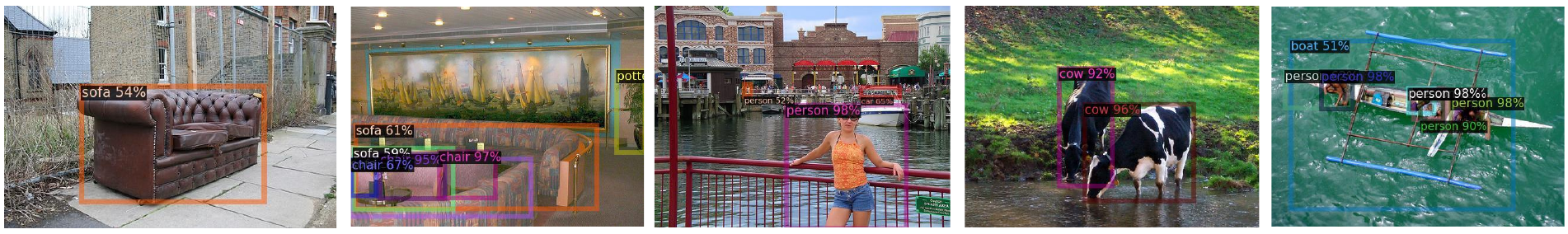}}
    \caption{The visualization results of DeFRCN, MFDC, and our approach are based on both of these methods. The models are under 2-shot on the Novel Split 3 setting.}
    \label{figvis}
\end{figure}

\noindent\textbf{Module pluggable verification.} \enspace In order to further validate the effectiveness of our proposed method, we conduct pluggable experiments on different baselines. Table \ref{tab:baseline} shows the performance of our method on TFA, DeFRCN, and MFDC at Novel Set 1. Our method has improved performance by 10.1\%, 6.0\%, and 0.9\% on three baselines, respectively. The overall performance shows that our method is pluggable and exhibits performance growth across various baselines.

\noindent\textbf{Visualization.} \enspace We provide some visualizations of PASCAL VOC datasets. As shown in Figure \ref{figvis}, our method can effectively reduce the occurrence of identifying confusion categories. In DeFRCN, there are many cases of missing detections while our method can detect these samples well. In MFDC, most samples can be detected, but error detections are prone to occur. Through our method, we can effectively rectify incorrect recognition.

\section{Conclusion}

In this work, we propose to \textbf{G}enerate Local \textbf{R}everse \textbf{S}amples as Few-shot Object \textbf{Det}ector (GRSDet), which can optimize the centers and boundaries of the novel class distributions in the decision space. It contains two plug-and-play modules, which are the Center Calibration Variance Augmentation (CCVA) module and Feature Density Boundary Optimization (FDBO) module. The CCVA module can calibrate the novel sample centers without relying on other datasets by generating the LRSamples, while the FDBO model can adaptively optimize distribution boundaries by reweighting the importance of edge samples. Finally, we demonstrate the effectiveness of our method on both the universal model DeFRCN and the state-of-the-art model MFDC.



%%       \citet{<label>} ==> Jones et al. [21]
%%       \citep{<label>} ==> [21]
%%

\bibliographystyle{elsarticle-num-names} 
\bibliography{bibfile}





% \end{thebibliography}
\end{document}

%% file: tables/distance.tex
\centering
    \resizebox{\textwidth}{!}{
  \begin{tabular}{c| c c c c c c}
    \toprule
    \multirow{2}*{\shortstack{Point set}}&  \multicolumn{6}{c}{Novel classes} \\
    & boat & cat & motorbike & sheep & sofa & mean\\
    \midrule
    
    % DeFRCN&\makebox[0.07\textwidth][c]{1}&\makebox[0.07\textwidth][c]{1}&\makebox[0.07\textwidth][c]{1}\\
    normal distance w/o. LRSamples &0.0622&0.0728&0.0731&0.0697&0.0744&0.0704\\
    distance w. LRSamples&\textbf{0.0647}&\textbf{0.0748}&\textbf{0.0732}&\textbf{0.0733}&\textbf{0.0751}&\textbf{0.0722}\\
    % DSDO&1-shot&\textbf{0.284}&\textbf{0.204}&\textbf{0.264}\\

    \bottomrule
  \end{tabular}}

%% file: tables/Gx.tex
\centering
\setlength{\tabcolsep}{4.5mm}{
  \begin{tabular}{c|c}
  \hline

  $\mathcal{G}(\cdot)$& Expression\\

  \hline
    sigmoid&$
\left\{ \begin{array}{c}
	2/\left( 1+e^{-loss\left( x \right)} \right) \ \ x\in \mathcal{I}\\
	2/\left( 1+e^{loss\left( x \right)} \right) \ \ x\in \mathcal{D}\\
\end{array} \right.$ \\\hline

    exp&$
\left\{ \begin{array}{c}
	 e^{loss\left( x \right)}  \ \ x\in \mathcal{I}\\
	 e^{-loss\left( x \right)}  \ \ x\in \mathcal{D}\\
\end{array} \right.$ \\

  \hline

    linear&$
    \left\{ \begin{array}{c}
    	 1+loss\left( x \right)  \ \ x\in \mathcal{I}\\
    	 1-loss\left( x \right)  \ \ x\in \mathcal{D}\\
    \end{array} \right.$ \\

    \hline

  \end{tabular}}

%% file: tables/pascal_voc.tex
\centering
\resizebox{\textwidth}{!}{

% \setlength{\tabcolsep}{3.5pt}
% \small
\begin{tabular}{@{}l|ccccc|ccccc|ccccc|c@{}}
\toprule
\multirow{2}{*}{Method / Shots}           & \multicolumn{5}{c|}{Novel Set 1}                                         & \multicolumn{5}{c|}{Novel Set 2}                                         & \multicolumn{5}{c|}{Novel Set 3}                                       & \multirow{2}{*}{Avg.}    \\ 
                                        & 1            & 2            & 3            & 5            & 10           & 1            & 2            & 3            & 5            & 10           & 1            & 2            & 3            & 5            & 10        &    \\ \midrule
FSRW~\cite{kang2019few}                            & 14.8         & 15.5         & 26.7         & 33.9         & 47.2         & 15.7         & 15.3         & 22.7         & 30.1         & 40.5         & 21.3         & 25.6         & 28.4         & 42.8         & 45.9      & 28.4   \\
MetaDet~\cite{wang2019meta}                     & 18.9         & 20.6         & 30.2         & 36.8         & 49.6         & 21.8         & 23.1         & 27.8         & 31.7         & 43.0         & 20.6         & 23.9         & 29.4         & 43.9         & 44.1      & 31.0   \\
Meta R-CNN~\cite{yan2019meta}                     & 19.9         & 25.5         & 35.0         & 45.7         & 51.5         & 10.4         & 19.4         & 29.6         & 34.8         & 45.4         & 14.3         & 18.2         & 27.5         & 41.2         & 48.1      & 31.1   \\

RepMet~\cite{karlinsky2019repmet}      &26.1&32.9&34.4&38.6&41.3&17.2&22.1&23.4&28.3&35.8&27.5&31.1&31.5&34.4&37.2&30.8\\

NP-RepMet~\cite{yang2020restoring}      &37.8&40.3&41.7&47.3&49.4&41.6&43.0&43.4&47.4&49.1&33.3&38.0&39.8&41.5&44.8&42.6\\

TFA w/ cos~\cite{wang2020frustratingly}           & 39.8         & 36.1         & 44.7         & 55.7         & 56.0         & 23.5         & 26.9         & 34.1         & 35.1         & 39.1         & 30.8         & 34.8         & 42.8         & 49.5         & 49.8      & 39.9   \\

HallucFsDet~\cite{zhang2021hallucination}                   & 47.0         & 44.9         & 46.5         & 54.7         & 54.7         & 26.3         & 31.8         & 37.4         & 37.4         & 41.2         & 40.4         & 42.1         & 43.3         & 51.4         & 49.6      & 43.2   \\
CGDP+FSCN~\cite{li2021few}                    & 40.7         & 45.1         & 46.5         & 57.4         & 62.4         & 27.3         & 31.4         & 40.8         & 42.7         & 46.3         & 31.2         & 36.4         & 43.7         & 50.1         & 55.6      & 43.8   \\
CME~\cite{li2021beyond}                     & 41.5         & 47.5         & 50.4         & 58.2         & 60.9         & 27.2         & 30.2         & 41.4         & 42.5         & 46.8         & 34.3         & 39.6         & 45.1         & 48.3         & 51.5      & 44.4   \\
SRR-FSD~\cite{zhu2021semantic}                 & 47.8         & 50.5         & 51.3         & 55.2         & 56.8         & 32.5         & 35.3         & 39.1         & 40.8         & 43.8         & 40.1         & 41.5         & 44.3         & 46.9         & 46.4      & 44.8  \\
UP-FSOD~\cite{wu2021universal}                   & 43.8         & 47.8         & 50.3         & 55.4         & 61.7         & 31.2         & 30.5         & 41.2         & 42.2         & 48.3         & 35.5         & 39.7         & 43.9         & 50.6         & 53.5      & 45.0   \\
FSCE~\cite{sun2021fsce}                & 44.2         & 43.8         & 51.4         & 61.9         & 63.4         & 27.3         & 29.5         & 43.5         & 44.2         & 50.2         & 37.2         & 41.9         & 47.5         & 54.6         & 58.5      & 46.6  \\

FADI~\cite{cao2021few}                      & 50.3         & 54.8         & 54.2         & 59.3         & 63.2         & 30.6         & 35.0         & 40.3         & 42.8         & 48.0         & 45.7         & 49.7         & 49.1         & 55.0         & 59.6   & 49.2   \\
SQMG~\cite{zhang2021accurate}               & 48.6         & 51.1         & 52.0         & 53.7         & 54.3         & 41.6   & 45.4    & 45.8         & 46.3         & 48.0         & 46.1         & 51.7   & 52.6         & 54.1         & 55.0      & 49.8   \\ 

EEKT~\cite{zhao2022exploring}  &46.7&53.1&53.8&61.0&62.1&30.1&34.2&41.6&41.9&44.8&41.0&46.0&47.2&55.4&55.6&47.6 \\

QA-FewDet~\cite{han2021query}&42.4&51.9&55.7&62.6&63.4&25.9&37.8&46.6&48.9&51.1&35.2&42.9&47.8&54.8&53.5&48.0\\ \midrule

DeFRCN~\cite{qiao2021defrcn}&57.0&58.6&64.3&67.8&67.0&35.8&42.7&51.0&54.5&52.9&52.5&56.6&55.8&60.7&62.5&56.0\\

% DSDO-DeFRCN&62.6&65.9&66.5&\textcolor{blue}{69.5}&\textcolor{red}{69.3}&41.5&\textcolor{blue}{47.2}&\textcolor{red}{53.5}&\textcolor{blue}{55.6}&\textcolor{red}{55.9}&\textcolor{blue}{57.9}&\textcolor{red}{60.5}&\textcolor{red}{60.6}&\textcolor{blue}{63.0}&\textcolor{blue}{64.3}&\textcolor{blue}{59.4}\\ \midrule

% MFDC~\cite{wu2022multi}&\textcolor{blue}{63.4}&\textcolor{blue}{66.3}&\textcolor{blue}{67.7}&69.4&68.1&\textcolor{blue}{42.1}&46.5&\textcolor{blue}{53.4}&55.3&\textcolor{blue}{53.8}&56.1&58.3&59.0&62.2&63.7&59.0\\

% DSDO-MFDC&\textcolor{red}{64.5}&\textcolor{red}{66.9}&\textcolor{red}{68.0}&\textcolor{red}{69.9}&\textcolor{blue}{68.2}&\textcolor{red}{42.7}&\textcolor{red}{47.6}&53.2&\textcolor{red}{55.4}&51.8&\textcolor{red}{58.4}&\textcolor{blue}{59.8}&\textcolor{blue}{60.5}&\textcolor{red}{63.3}&\textcolor{red}{65.3}&\textcolor{red}{59.7}\\

GRSDet-DeFRCN&62.6&65.9&66.5&69.5&\textbf{69.3}&41.5&47.2&\textbf{53.5}&\textbf{55.6}&\textbf{55.9}& 57.9&\textbf{60.5}&\textbf{60.6}& 63.0& 64.3& \textbf{59.4}\\ \midrule

MFDC~\cite{wu2022multi}& 63.4& 66.3& 67.7&69.4&68.1& 42.1&46.5& 53.4&55.3& 53.8&56.1&58.3&59.0&62.2&63.7&59.0\\

GRSDet-MFDC&\textbf{64.8}&\textbf{66.6}&\textbf{68.0}&\textbf{70.0}& 68.2&\textbf{42.7}&\textbf{47.6}&53.2&55.4&51.8&\textbf{58.2}& 59.8& 60.5&\textbf{63.3}&\textbf{65.3}&\textbf{59.7}\\

\bottomrule
\end{tabular}}

%% file: tables/ms_coco.tex
\centering
\resizebox{\textwidth}{!}{
    % \scriptsize
  \begin{tabular}{l| c c | c c | c c | c c | c c | c c}
    \toprule
    \multirow{2}{*}{Method}  & \multicolumn{2}{c|}{1-shot} & \multicolumn{2}{c|}{2-shot} & \multicolumn{2}{c|}{3-shot} & \multicolumn{2}{c|}{5-shot} & \multicolumn{2}{c|}{10-shot} & \multicolumn{2}{c}{30-shot}\\
    & nAP & nAP75 & nAP & nAP75 & nAP & nAP75 & nAP & nAP75 & nAP & nAP75 & nAP & nAP75\\
    \midrule
    % -&&&&&&&&&&&&\\
    FSRW~\cite{kang2019few}&$-$&$-$&$-$&$-$&$-$&$-$&$-$&$-$&5.6&4.6&9.1&7.6\\
    SRR-FSD~\cite{zhu2021semantic}&$-$&$-$&$-$&$-$&$-$&$-$&$-$&$-$&11.3&9.8&14.7&13.5\\
    FSCE~\cite{sun2021fsce}&$-$&$-$&$-$&$-$&$-$&$-$&$-$&$-$&11.9&10.5&16.4&16.2\\
    UP-FSOD~\cite{wu2021universal}&$-$&$-$&$-$&$-$&$-$&$-$&$-$&$-$&11.0&10.7&15.6&15.7\\
    SQMG~\cite{zhang2021accurate}&$-$&$-$&$-$&$-$&$-$&$-$&$-$&$-$&13.9&11.7&15.9&14.3\\
    CME~\cite{li2021beyond}&$-$&$-$&$-$&$-$&$-$&$-$&$-$&$-$&15.1&16.4&16.9&17.8\\
    TFA w/cos~\cite{wang2020frustratingly}&3.4&3.8&4.6&4.8&6.6&6.5&8.3&8.0&10.0&9.3&13.7&13.4\\
    
    QA-FewDet~\cite{han2021query}&4.9&4.4&7.6&6.2&8.4&7.3&9.7&8.6&11.6&9.8&16.5&15.5\\
    FADI~\cite{cao2021few}&5.7&6.0&7.0&7.0&8.6&8.3&10.1&9.7&12.2&11.9&16.1&15.8\\
    
    \midrule
    DeFRCN~\cite{qiao2021defrcn}&6.5&6.9&11.8&12.4&13.4&13.6&15.3&14.6&18.6&17.6&22.5&22.3\\
    GRSDet-DeFRCN&9.0&9.4&12.1&12.5&14.3&14.1&15.7&15.5&18.5&18.6&22.6&\textbf{23.4}\\
    
    \midrule
    MFDC~\cite{wu2022multi}&10.8&11.6&13.9&14.8&15.0&15.5&16.4&17.3&19.4&\textbf{20.2}&\textbf{22.7}&23.2\\
    GRSDet-MFDC&\textbf{11.5}&\textbf{12.6}&\textbf{14.3}&\textbf{14.8}&\textbf{16.0}&\textbf{16.6}&\textbf{17.1}&\textbf{17.4}&\textbf{19.5}&20.0&22.4&23.0
    \\
    \bottomrule
  \end{tabular}}

%% file: tables/ablation.tex
\centering
    % \scriptsize
  \begin{tabular}{c c c| c c c}
    \toprule
    \multirow{2}*{\shortstack{CCVA}}& \multirow{2}*{\shortstack{FDBO-$\mathcal{I}$}}& \multirow{2}*{\shortstack{FDBO-$\mathcal{D}$}} &  \multicolumn{3}{c}{nAP50} \\
    & & & 1-shot & 2-shot & 3-shot\\
    \midrule
    &&&52.3&58.7&58.6\\
    $\checkmark$&&&57.0&60.3&59.6\\
    &$\checkmark$&&54.8&59.1&59.1\\
    &&$\checkmark$&54.6&58.9&60.2 \\
    $\checkmark$ & & $\checkmark$ &57.7  & 60.0 & 60.6  \\
    $\checkmark$&$\checkmark$&& 57.2 & 60.2 & 59.7 \\
    &$\checkmark$&$\checkmark$& 54.6 &59.7& 60.3 \\
    $\checkmark$ & $\checkmark$ & $\checkmark$ &\textbf{57.9}&\textbf{60.5}&\textbf{60.6}\\
    \bottomrule
  \end{tabular}

%% file: tables/CA.tex
\centering
    % \scriptsize
  \begin{tabular}{c | c c c c c}
    \toprule
    \multirow{2}*{Center Calibration} &  \multicolumn{3}{c}{nAP50} \\
    & 1-shot & 2-shot & 3-shot & 5-shot & 10-shot\\
    \midrule
    w/o.&56.6&  58.6 & 59.2 & 62.0 &63.2\\
    w/.&\textbf{57.0}&\textbf{60.3}&\textbf{59.6}&\textbf{62.5}&\textbf{64.3}\\
    \bottomrule
  \end{tabular}

%% file: tables/base.tex
\centering
    % \scriptsize
  \begin{tabular}{c | c c c c c}
    \toprule
    \multirow{2}*{Method} &  \multicolumn{5}{c}{bAP50} \\
    & 1-shot& 2-shot & 3-shot & 5-shot & 10-shot\\
    \midrule
    MFDC&79.79&79.96&\textbf{80.07}&79.98&79.58\\
    DSDO&\textbf{80.34}&\textbf{80.28}&79.72&\textbf{80.40}&\textbf{79.89}\\
    \bottomrule
  \end{tabular}

%% file: tables/loss_base.tex
\centering
    % \scriptsize
  \begin{tabular}{c | c | c c c c c}
    \toprule
    \multirow{2}*{$\lambda_{1}$} & \multirow{2}*{$\lambda_{2}$} &  \multicolumn{5}{c}{nAP50} \\
      &  & 1-shot& 2-shot & 3-shot & 5-shot & 10-shot\\
    \midrule
    0.03&0.4&56.4&59.6&58.0&61.7&63.1\\
    0.1&0.4&57.4&58.6&58.4&61.6&61.5\\
    0.05&0.4&\textbf{57.9}&\textbf{60.5}&\textbf{60.6}&\textbf{63.0}&\textbf{64.3}\\
    0.05&0.2&53.5&57.4&56.3&61.1&63.8\\
    0.05&0.6&54.9&59.0&57.9&61.7&63.5\\
    \bottomrule
  \end{tabular}

%% file: tables/loss_finetune.tex
\centering
    % \scriptsize
  \begin{tabular}{c | c | c c c c c}
    \toprule
    \multirow{2}*{$\lambda_{3}$} & \multirow{2}*{$\lambda_{4}$} &  \multicolumn{5}{c}{nAP50} \\
      &  & 1-shot& 2-shot & 3-shot & 5-shot & 10-shot\\
    \midrule
    0.1&0.1&55.7&59.9&58.5&62.9&\textbf{64.5}\\
    0.5&0.1&55.2&57.8&56.5&61.0&63.3\\
    0.3&0.1&\textbf{57.9}&\textbf{60.5}&\textbf{60.6}&\textbf{63.0}&64.3\\
    0.3&0.05&55.0&58.9&57.7&62.1&63.2\\
    0.3&0.2&55.8&58.3&57.6&62.1&63.5\\
    \bottomrule
  \end{tabular}

%% file: tables/ifc.tex
\centering
    % \scriptsize
  \begin{tabular}{c | c c c c c}
    \toprule
    \multirow{2}*{Structure} &  \multicolumn{5}{c}{nAP50} \\
    & 1-shot& 2-shot & 3-shot & 5-shot & 10-shot\\
    \midrule
    Low channel&65.2&61.1&67.6&68.0&67.5\\
    High channel&64.6&\textbf{67.9}&\textbf{69.1}&69.6&67.3\\
    LeakyReLu&64.1&64.2&67.1&68.9&67.5\\
    Equal channel&\textbf{64.8}&66.6&68.0&\textbf{70.0}&\textbf{68.2}\\
    \bottomrule
  \end{tabular}

%% file: tables/baseline.tex
\centering
    % \scriptsize
  \begin{tabular}{c | c | c c c c c | c}
    \toprule
    \multirow{2}{*}{Method} & \multirow{2}{*}{DSDO} & \multicolumn{5}{c|}{nAP50} & \multirow{2}{*}{Avg.} \\
    & & 1-shot& 2-shot & 3-shot & 5-shot & 10-shot&\\
    \midrule
    \multirow{2}*{TFA w/ cos}& &39.8&36.1&44.7&55.7&56.0&46.5\\
    & \checkmark&46.5&42.4&51.3&57.8&58.0&51.2\\
    \midrule
    \multirow{2}*{DeFRCN}& & 57.0&58.6&64.3&67.8&67.0&62.9\\
     & \checkmark&62.6&65.9&66.5&69.5&69.3&66.7\\
    \midrule
    \multirow{2}*{MFDC}& & 63.4& 66.3& 67.7&69.4&68.1&66.9\\
    & \checkmark&64.8&66.6&68.0&70.0&68.2&67.5\\
    \bottomrule
  \end{tabular}

%% file: elsarticle-template-num-names.bbl
\begin{thebibliography}{59}
\expandafter\ifx\csname natexlab\endcsname\relax\def\natexlab#1{#1}\fi
\providecommand{\url}[1]{\texttt{#1}}
\providecommand{\href}[2]{#2}
\providecommand{\path}[1]{#1}
\providecommand{\DOIprefix}{doi:}
\providecommand{\ArXivprefix}{arXiv:}
\providecommand{\URLprefix}{URL: }
\providecommand{\Pubmedprefix}{pmid:}
\providecommand{\doi}[1]{\href{http://dx.doi.org/#1}{\path{#1}}}
\providecommand{\Pubmed}[1]{\href{pmid:#1}{\path{#1}}}
\providecommand{\bibinfo}[2]{#2}
\ifx\xfnm\relax \def\xfnm[#1]{\unskip,\space#1}\fi
%Type = Article
\bibitem[{Wang et~al.(2020)Wang, Huang, Darrell, Gonzalez, and
  Yu}]{wang2020frustratingly}
\bibinfo{author}{X.~Wang}, \bibinfo{author}{T.~E. Huang},
  \bibinfo{author}{T.~Darrell}, \bibinfo{author}{J.~E. Gonzalez},
  \bibinfo{author}{F.~Yu},
\newblock \bibinfo{title}{Frustratingly simple few-shot object detection},
\newblock \bibinfo{journal}{Proceedings of the international conference on
  machine learning}  (\bibinfo{year}{2020}) \bibinfo{pages}{9919--9928}.
%Type = Inproceedings
\bibitem[{Yan et~al.(2019)Yan, Chen, Xu, Wang, Liang, and Lin}]{yan2019meta}
\bibinfo{author}{X.~Yan}, \bibinfo{author}{Z.~Chen}, \bibinfo{author}{A.~Xu},
  \bibinfo{author}{X.~Wang}, \bibinfo{author}{X.~Liang},
  \bibinfo{author}{L.~Lin},
\newblock \bibinfo{title}{Meta r-cnn: Towards general solver for instance-level
  low-shot learning},
\newblock in: \bibinfo{booktitle}{Proceedings of the IEEE international
  conference on computer vision}, \bibinfo{year}{2019}, pp.
  \bibinfo{pages}{9577--9586}.
%Type = Inproceedings
\bibitem[{Karlinsky et~al.(2019)Karlinsky, Shtok, Harary, Schwartz, Aides,
  Feris, Giryes, and Bronstein}]{karlinsky2019repmet}
\bibinfo{author}{L.~Karlinsky}, \bibinfo{author}{J.~Shtok},
  \bibinfo{author}{S.~Harary}, \bibinfo{author}{E.~Schwartz},
  \bibinfo{author}{A.~Aides}, \bibinfo{author}{R.~Feris},
  \bibinfo{author}{R.~Giryes}, \bibinfo{author}{A.~M. Bronstein},
\newblock \bibinfo{title}{Repmet: Representative-based metric learning for
  classification and few-shot object detection},
\newblock in: \bibinfo{booktitle}{Proceedings of the IEEE conference on
  computer vision and pattern recognition}, \bibinfo{year}{2019}, pp.
  \bibinfo{pages}{5197--5206}.
%Type = Inproceedings
\bibitem[{Qiao et~al.(2021)Qiao, Zhao, Li, Qiu, Wu, and Zhang}]{qiao2021defrcn}
\bibinfo{author}{L.~Qiao}, \bibinfo{author}{Y.~Zhao}, \bibinfo{author}{Z.~Li},
  \bibinfo{author}{X.~Qiu}, \bibinfo{author}{J.~Wu},
  \bibinfo{author}{C.~Zhang},
\newblock \bibinfo{title}{Defrcn: Decoupled faster r-cnn for few-shot object
  detection},
\newblock in: \bibinfo{booktitle}{Proceedings of the IEEE international
  conference on computer vision}, \bibinfo{year}{2021}, pp.
  \bibinfo{pages}{8681--8690}.
%Type = Article
\bibitem[{Leng et~al.(2022)Leng, Chen, Gao, Mo, Yu, and
  Zhang}]{leng2022sampling}
\bibinfo{author}{J.~Leng}, \bibinfo{author}{T.~Chen}, \bibinfo{author}{X.~Gao},
  \bibinfo{author}{M.~Mo}, \bibinfo{author}{Y.~Yu}, \bibinfo{author}{Y.~Zhang},
\newblock \bibinfo{title}{Sampling-invariant fully metric learning for few-shot
  object detection},
\newblock \bibinfo{journal}{Neurocomputing} \bibinfo{volume}{511}
  (\bibinfo{year}{2022}) \bibinfo{pages}{54--66}.
%Type = Article
\bibitem[{Du et~al.(2022)Du, Liu, Jiao, Hao, Li, Liu, and
  Liu}]{du2022augmentative}
\bibinfo{author}{Y.~Du}, \bibinfo{author}{F.~Liu}, \bibinfo{author}{L.~Jiao},
  \bibinfo{author}{Z.~Hao}, \bibinfo{author}{S.~Li}, \bibinfo{author}{X.~Liu},
  \bibinfo{author}{J.~Liu},
\newblock \bibinfo{title}{Augmentative contrastive learning for one-shot object
  detection},
\newblock \bibinfo{journal}{Neurocomputing} \bibinfo{volume}{513}
  (\bibinfo{year}{2022}) \bibinfo{pages}{13--24}.
%Type = Inproceedings
\bibitem[{Sun et~al.(2021)Sun, Li, Cai, Yuan, and Zhang}]{sun2021fsce}
\bibinfo{author}{B.~Sun}, \bibinfo{author}{B.~Li}, \bibinfo{author}{S.~Cai},
  \bibinfo{author}{Y.~Yuan}, \bibinfo{author}{C.~Zhang},
\newblock \bibinfo{title}{Fsce: Few-shot object detection via contrastive
  proposal encoding},
\newblock in: \bibinfo{booktitle}{Proceedings of the IEEE conference on
  computer vision and pattern recognition}, \bibinfo{year}{2021}, pp.
  \bibinfo{pages}{7352--7362}.
%Type = Article
\bibitem[{Zhang et~al.(2022)Zhang, Luo, Cui, Lu, and Xing}]{zhang2022meta}
\bibinfo{author}{G.~Zhang}, \bibinfo{author}{Z.~Luo}, \bibinfo{author}{K.~Cui},
  \bibinfo{author}{S.~Lu}, \bibinfo{author}{E.~P. Xing},
\newblock \bibinfo{title}{Meta-detr: Image-level few-shot detection with
  inter-class correlation exploitation},
\newblock \bibinfo{journal}{IEEE Transactions on Pattern Analysis and Machine
  Intelligence}  (\bibinfo{year}{2022}).
%Type = Article
\bibitem[{Cao et~al.(2021)Cao, Wang, Jin, Wu, Chen, Liu, and Lin}]{cao2021few}
\bibinfo{author}{Y.~Cao}, \bibinfo{author}{J.~Wang}, \bibinfo{author}{Y.~Jin},
  \bibinfo{author}{T.~Wu}, \bibinfo{author}{K.~Chen}, \bibinfo{author}{Z.~Liu},
  \bibinfo{author}{D.~Lin},
\newblock \bibinfo{title}{Few-shot object detection via association and
  discrimination},
\newblock \bibinfo{journal}{Advances in neural information processing systems}
  \bibinfo{volume}{34} (\bibinfo{year}{2021}) \bibinfo{pages}{16570--16581}.
%Type = Inproceedings
\bibitem[{Zhao et~al.(2022)Zhao, Liu, and Wang}]{zhao2022exploring}
\bibinfo{author}{Z.~Zhao}, \bibinfo{author}{Q.~Liu}, \bibinfo{author}{Y.~Wang},
\newblock \bibinfo{title}{Exploring effective knowledge transfer for few-shot
  object detection},
\newblock in: \bibinfo{booktitle}{Proceedings of the 30th ACM International
  Conference on Multimedia}, \bibinfo{year}{2022}, pp.
  \bibinfo{pages}{6831--6839}.
%Type = Inproceedings
\bibitem[{Deng et~al.(2009)Deng, Dong, Socher, Li, Li, and
  Fei-Fei}]{deng2009imagenet}
\bibinfo{author}{J.~Deng}, \bibinfo{author}{W.~Dong},
  \bibinfo{author}{R.~Socher}, \bibinfo{author}{L.-J. Li},
  \bibinfo{author}{K.~Li}, \bibinfo{author}{L.~Fei-Fei},
\newblock \bibinfo{title}{Imagenet: A large-scale hierarchical image database},
\newblock in: \bibinfo{booktitle}{2009 IEEE conference on computer vision and
  pattern recognition}, \bibinfo{organization}{Ieee}, \bibinfo{year}{2009}, pp.
  \bibinfo{pages}{248--255}.
%Type = Article
\bibitem[{Miller(1995)}]{miller1995wordnet}
\bibinfo{author}{G.~A. Miller},
\newblock \bibinfo{title}{Wordnet: a lexical database for english},
\newblock \bibinfo{journal}{Communications of the ACM} \bibinfo{volume}{38}
  (\bibinfo{year}{1995}) \bibinfo{pages}{39--41}.
%Type = Inproceedings
\bibitem[{Wu et~al.(2022)Wu, Pei, Mei, Chen, Tian, and Lu}]{wu2022multi}
\bibinfo{author}{S.~Wu}, \bibinfo{author}{W.~Pei}, \bibinfo{author}{D.~Mei},
  \bibinfo{author}{F.~Chen}, \bibinfo{author}{J.~Tian},
  \bibinfo{author}{G.~Lu},
\newblock \bibinfo{title}{Multi-faceted distillation of base-novel commonality
  for few-shot object detection},
\newblock in: \bibinfo{booktitle}{Proceedings of the European conference on
  computer vision}, \bibinfo{organization}{Springer}, \bibinfo{year}{2022}, pp.
  \bibinfo{pages}{578--594}.
%Type = Article
\bibitem[{Everingham et~al.(2010)Everingham, Van~Gool, Williams, Winn, and
  Zisserman}]{everingham2010pascal}
\bibinfo{author}{M.~Everingham}, \bibinfo{author}{L.~Van~Gool},
  \bibinfo{author}{C.~K. Williams}, \bibinfo{author}{J.~Winn},
  \bibinfo{author}{A.~Zisserman},
\newblock \bibinfo{title}{The pascal visual object classes (voc) challenge},
\newblock \bibinfo{journal}{International journal of computer vision}
  \bibinfo{volume}{88} (\bibinfo{year}{2010}) \bibinfo{pages}{303--338}.
%Type = Inproceedings
\bibitem[{Lin et~al.(2014)Lin, Maire, Belongie, Hays, Perona, Ramanan,
  Doll{\'a}r, and Zitnick}]{lin2014microsoft}
\bibinfo{author}{T.-Y. Lin}, \bibinfo{author}{M.~Maire},
  \bibinfo{author}{S.~Belongie}, \bibinfo{author}{J.~Hays},
  \bibinfo{author}{P.~Perona}, \bibinfo{author}{D.~Ramanan},
  \bibinfo{author}{P.~Doll{\'a}r}, \bibinfo{author}{C.~L. Zitnick},
\newblock \bibinfo{title}{Microsoft coco: Common objects in context},
\newblock in: \bibinfo{booktitle}{Proceedings of the European conference on
  computer vision}, \bibinfo{organization}{Springer}, \bibinfo{year}{2014}, pp.
  \bibinfo{pages}{740--755}.
%Type = Inproceedings
\bibitem[{Jamal and Qi(2019)}]{jamal2019task}
\bibinfo{author}{M.~A. Jamal}, \bibinfo{author}{G.-J. Qi},
\newblock \bibinfo{title}{Task agnostic meta-learning for few-shot learning},
\newblock in: \bibinfo{booktitle}{Proceedings of the IEEE conference on
  computer vision and pattern recognition}, \bibinfo{year}{2019}, pp.
  \bibinfo{pages}{11719--11727}.
%Type = Inproceedings
\bibitem[{Finn et~al.(2017)Finn, Abbeel, and Levine}]{finn2017model}
\bibinfo{author}{C.~Finn}, \bibinfo{author}{P.~Abbeel},
  \bibinfo{author}{S.~Levine},
\newblock \bibinfo{title}{Model-agnostic meta-learning for fast adaptation of
  deep networks},
\newblock in: \bibinfo{booktitle}{International conference on machine
  learning}, \bibinfo{organization}{PMLR}, \bibinfo{year}{2017}, pp.
  \bibinfo{pages}{1126--1135}.
%Type = Inproceedings
\bibitem[{Naik and Mammone(1992)}]{naik1992meta}
\bibinfo{author}{D.~K. Naik}, \bibinfo{author}{R.~J. Mammone},
\newblock \bibinfo{title}{Meta-neural networks that learn by learning},
\newblock in: \bibinfo{booktitle}{[Proceedings 1992] IJCNN International Joint
  Conference on Neural Networks}, volume~\bibinfo{volume}{1},
  \bibinfo{organization}{IEEE}, \bibinfo{year}{1992}, pp.
  \bibinfo{pages}{437--442}.
%Type = Article
\bibitem[{Bertinetto et~al.(2018)Bertinetto, Henriques, Torr, and
  Vedaldi}]{bertinetto2018meta}
\bibinfo{author}{L.~Bertinetto}, \bibinfo{author}{J.~F. Henriques},
  \bibinfo{author}{P.~H. Torr}, \bibinfo{author}{A.~Vedaldi},
\newblock \bibinfo{title}{Meta-learning with differentiable closed-form
  solvers},
\newblock \bibinfo{journal}{International Conference on Learning
  Representations}  (\bibinfo{year}{2018}).
%Type = Inproceedings
\bibitem[{Simon et~al.(2020)Simon, Koniusz, Nock, and
  Harandi}]{simon2020modulating}
\bibinfo{author}{C.~Simon}, \bibinfo{author}{P.~Koniusz},
  \bibinfo{author}{R.~Nock}, \bibinfo{author}{M.~Harandi},
\newblock \bibinfo{title}{On modulating the gradient for meta-learning},
\newblock in: \bibinfo{booktitle}{Proceedings of the European conference on
  computer vision}, \bibinfo{organization}{Springer}, \bibinfo{year}{2020}, pp.
  \bibinfo{pages}{556--572}.
%Type = Article
\bibitem[{Zhu and Li(2022)}]{zhu2022mgml}
\bibinfo{author}{X.~Zhu}, \bibinfo{author}{S.~Li},
\newblock \bibinfo{title}{Mgml: Momentum group meta-learning for few-shot image
  classification},
\newblock \bibinfo{journal}{Neurocomputing} \bibinfo{volume}{514}
  (\bibinfo{year}{2022}) \bibinfo{pages}{351--361}.
%Type = Inproceedings
\bibitem[{Sung et~al.(2018)Sung, Yang, Zhang, Xiang, Torr, and
  Hospedales}]{sung2018learning}
\bibinfo{author}{F.~Sung}, \bibinfo{author}{Y.~Yang},
  \bibinfo{author}{L.~Zhang}, \bibinfo{author}{T.~Xiang},
  \bibinfo{author}{P.~H. Torr}, \bibinfo{author}{T.~M. Hospedales},
\newblock \bibinfo{title}{Learning to compare: Relation network for few-shot
  learning},
\newblock in: \bibinfo{booktitle}{Proceedings of the IEEE conference on
  computer vision and pattern recognition}, \bibinfo{year}{2018}, pp.
  \bibinfo{pages}{1199--1208}.
%Type = Article
\bibitem[{Snell et~al.(2017)Snell, Swersky, and Zemel}]{snell2017prototypical}
\bibinfo{author}{J.~Snell}, \bibinfo{author}{K.~Swersky},
  \bibinfo{author}{R.~Zemel},
\newblock \bibinfo{title}{Prototypical networks for few-shot learning},
\newblock \bibinfo{journal}{Advances in neural information processing systems}
  \bibinfo{volume}{30} (\bibinfo{year}{2017}).
%Type = Article
\bibitem[{Vinyals et~al.(2016)Vinyals, Blundell, Lillicrap, Wierstra
  et~al.}]{vinyals2016matching}
\bibinfo{author}{O.~Vinyals}, \bibinfo{author}{C.~Blundell},
  \bibinfo{author}{T.~Lillicrap}, \bibinfo{author}{D.~Wierstra}, et~al.,
\newblock \bibinfo{title}{Matching networks for one shot learning},
\newblock \bibinfo{journal}{Advances in neural information processing systems}
  \bibinfo{volume}{29} (\bibinfo{year}{2016}).
%Type = Article
\bibitem[{Garcia and Bruna(2018)}]{garcia2017few}
\bibinfo{author}{V.~Garcia}, \bibinfo{author}{J.~Bruna},
\newblock \bibinfo{title}{Few-shot learning with graph neural networks},
\newblock \bibinfo{journal}{International Conference on Learning
  Representations}  (\bibinfo{year}{2018}).
%Type = Inproceedings
\bibitem[{Zhang et~al.(2020)Zhang, Cai, Lin, and Shen}]{zhang2020deepemd}
\bibinfo{author}{C.~Zhang}, \bibinfo{author}{Y.~Cai}, \bibinfo{author}{G.~Lin},
  \bibinfo{author}{C.~Shen},
\newblock \bibinfo{title}{Deepemd: Few-shot image classification with
  differentiable earth mover's distance and structured classifiers},
\newblock in: \bibinfo{booktitle}{Proceedings of the IEEE conference on
  computer vision and pattern recognition}, \bibinfo{year}{2020}, pp.
  \bibinfo{pages}{12203--12213}.
%Type = Article
\bibitem[{Liu et~al.(2018)Liu, Lee, Park, Kim, Yang, Hwang, and
  Yang}]{liu2018learning}
\bibinfo{author}{Y.~Liu}, \bibinfo{author}{J.~Lee}, \bibinfo{author}{M.~Park},
  \bibinfo{author}{S.~Kim}, \bibinfo{author}{E.~Yang}, \bibinfo{author}{S.~J.
  Hwang}, \bibinfo{author}{Y.~Yang},
\newblock \bibinfo{title}{Learning to propagate labels: Transductive
  propagation network for few-shot learning},
\newblock \bibinfo{journal}{arXiv preprint arXiv:1805.10002}
  (\bibinfo{year}{2018}).
%Type = Inproceedings
\bibitem[{Li et~al.(2020)Li, Zhang, Li, and Fu}]{li2020adversarial}
\bibinfo{author}{K.~Li}, \bibinfo{author}{Y.~Zhang}, \bibinfo{author}{K.~Li},
  \bibinfo{author}{Y.~Fu},
\newblock \bibinfo{title}{Adversarial feature hallucination networks for
  few-shot learning},
\newblock in: \bibinfo{booktitle}{Proceedings of the IEEE conference on
  computer vision and pattern recognition}, \bibinfo{year}{2020}, pp.
  \bibinfo{pages}{13470--13479}.
%Type = Inproceedings
\bibitem[{Li et~al.(2019)Li, Xu, Huo, Wang, Gao, and Luo}]{li2019distribution}
\bibinfo{author}{W.~Li}, \bibinfo{author}{J.~Xu}, \bibinfo{author}{J.~Huo},
  \bibinfo{author}{L.~Wang}, \bibinfo{author}{Y.~Gao},
  \bibinfo{author}{J.~Luo},
\newblock \bibinfo{title}{Distribution consistency based covariance metric
  networks for few-shot learning},
\newblock in: \bibinfo{booktitle}{Proceedings of the AAAI conference on
  artificial intelligence}, volume~\bibinfo{volume}{33}, \bibinfo{year}{2019},
  pp. \bibinfo{pages}{8642--8649}.
%Type = Article
\bibitem[{Oreshkin et~al.(2018)Oreshkin, Rodr{\'\i}guez~L{\'o}pez, and
  Lacoste}]{oreshkin2018tadam}
\bibinfo{author}{B.~Oreshkin}, \bibinfo{author}{P.~Rodr{\'\i}guez~L{\'o}pez},
  \bibinfo{author}{A.~Lacoste},
\newblock \bibinfo{title}{Tadam: Task dependent adaptive metric for improved
  few-shot learning},
\newblock \bibinfo{journal}{Advances in neural information processing systems}
  \bibinfo{volume}{31} (\bibinfo{year}{2018}).
%Type = Inproceedings
\bibitem[{Wang et~al.(2018)Wang, Girshick, Hebert, and Hariharan}]{wang2018low}
\bibinfo{author}{Y.-X. Wang}, \bibinfo{author}{R.~Girshick},
  \bibinfo{author}{M.~Hebert}, \bibinfo{author}{B.~Hariharan},
\newblock \bibinfo{title}{Low-shot learning from imaginary data},
\newblock in: \bibinfo{booktitle}{Proceedings of the IEEE conference on
  computer vision and pattern recognition}, \bibinfo{year}{2018}, pp.
  \bibinfo{pages}{7278--7286}.
%Type = Inproceedings
\bibitem[{Hariharan and Girshick(2017)}]{hariharan2017low}
\bibinfo{author}{B.~Hariharan}, \bibinfo{author}{R.~Girshick},
\newblock \bibinfo{title}{Low-shot visual recognition by shrinking and
  hallucinating features},
\newblock in: \bibinfo{booktitle}{Proceedings of the IEEE international
  conference on computer vision}, \bibinfo{year}{2017}, pp.
  \bibinfo{pages}{3018--3027}.
%Type = Inproceedings
\bibitem[{Xu et~al.(2021)Xu, Fu, Liu, Wang, Li, Huang, Zhang, and
  Xue}]{xu2021learning}
\bibinfo{author}{C.~Xu}, \bibinfo{author}{Y.~Fu}, \bibinfo{author}{C.~Liu},
  \bibinfo{author}{C.~Wang}, \bibinfo{author}{J.~Li},
  \bibinfo{author}{F.~Huang}, \bibinfo{author}{L.~Zhang},
  \bibinfo{author}{X.~Xue},
\newblock \bibinfo{title}{Learning dynamic alignment via meta-filter for
  few-shot learning},
\newblock in: \bibinfo{booktitle}{Proceedings of the IEEE conference on
  computer vision and pattern recognition}, \bibinfo{year}{2021}, pp.
  \bibinfo{pages}{5182--5191}.
%Type = Inproceedings
\bibitem[{Ma et~al.(2022)Ma, Fang, Avraham, Zuo, Zhu, Drummond, and
  Harandi}]{ma2022learning}
\bibinfo{author}{R.~Ma}, \bibinfo{author}{P.~Fang},
  \bibinfo{author}{G.~Avraham}, \bibinfo{author}{Y.~Zuo},
  \bibinfo{author}{T.~Zhu}, \bibinfo{author}{T.~Drummond},
  \bibinfo{author}{M.~Harandi},
\newblock \bibinfo{title}{Learning instance and task-aware dynamic kernels for
  few-shot learning},
\newblock in: \bibinfo{booktitle}{Proceedings of the European conference on
  computer vision}, \bibinfo{organization}{Springer}, \bibinfo{year}{2022}, pp.
  \bibinfo{pages}{257--274}.
%Type = Inproceedings
\bibitem[{Kang et~al.(2019)Kang, Liu, Wang, Yu, Feng, and
  Darrell}]{kang2019few}
\bibinfo{author}{B.~Kang}, \bibinfo{author}{Z.~Liu}, \bibinfo{author}{X.~Wang},
  \bibinfo{author}{F.~Yu}, \bibinfo{author}{J.~Feng},
  \bibinfo{author}{T.~Darrell},
\newblock \bibinfo{title}{Few-shot object detection via feature reweighting},
\newblock in: \bibinfo{booktitle}{Proceedings of the IEEE international
  conference on computer vision}, \bibinfo{year}{2019}, pp.
  \bibinfo{pages}{8420--8429}.
%Type = Inproceedings
\bibitem[{Fan et~al.(2020)Fan, Zhuo, Tang, and Tai}]{fan2020few}
\bibinfo{author}{Q.~Fan}, \bibinfo{author}{W.~Zhuo}, \bibinfo{author}{C.-K.
  Tang}, \bibinfo{author}{Y.-W. Tai},
\newblock \bibinfo{title}{Few-shot object detection with attention-rpn and
  multi-relation detector},
\newblock in: \bibinfo{booktitle}{Proceedings of the IEEE conference on
  computer vision and pattern recognition}, \bibinfo{year}{2020}, pp.
  \bibinfo{pages}{4013--4022}.
%Type = Inproceedings
\bibitem[{Hu et~al.(2021)Hu, Bai, Li, Cui, and Wang}]{hu2021dense}
\bibinfo{author}{H.~Hu}, \bibinfo{author}{S.~Bai}, \bibinfo{author}{A.~Li},
  \bibinfo{author}{J.~Cui}, \bibinfo{author}{L.~Wang},
\newblock \bibinfo{title}{Dense relation distillation with context-aware
  aggregation for few-shot object detection},
\newblock in: \bibinfo{booktitle}{Proceedings of the IEEE conference on
  computer vision and pattern recognition}, \bibinfo{year}{2021}, pp.
  \bibinfo{pages}{10185--10194}.
%Type = Inproceedings
\bibitem[{Zhang et~al.(2021)Zhang, Zhou, Guan, and Zhang}]{zhang2021accurate}
\bibinfo{author}{L.~Zhang}, \bibinfo{author}{S.~Zhou},
  \bibinfo{author}{J.~Guan}, \bibinfo{author}{J.~Zhang},
\newblock \bibinfo{title}{Accurate few-shot object detection with support-query
  mutual guidance and hybrid loss},
\newblock in: \bibinfo{booktitle}{Proceedings of the IEEE conference on
  computer vision and pattern recognition}, \bibinfo{year}{2021}, pp.
  \bibinfo{pages}{14424--14432}.
%Type = Inproceedings
\bibitem[{Wang et~al.(2019)Wang, Ramanan, and Hebert}]{wang2019meta}
\bibinfo{author}{Y.-X. Wang}, \bibinfo{author}{D.~Ramanan},
  \bibinfo{author}{M.~Hebert},
\newblock \bibinfo{title}{Meta-learning to detect rare objects},
\newblock in: \bibinfo{booktitle}{Proceedings of the IEEE international
  conference on computer vision}, \bibinfo{year}{2019}, pp.
  \bibinfo{pages}{9925--9934}.
%Type = Inproceedings
\bibitem[{Han et~al.(2022)Han, Huang, Ma, He, and Chang}]{han2022meta}
\bibinfo{author}{G.~Han}, \bibinfo{author}{S.~Huang}, \bibinfo{author}{J.~Ma},
  \bibinfo{author}{Y.~He}, \bibinfo{author}{S.-F. Chang},
\newblock \bibinfo{title}{Meta faster r-cnn: Towards accurate few-shot object
  detection with attentive feature alignment},
\newblock in: \bibinfo{booktitle}{Proceedings of the AAAI conference on
  artificial intelligence}, volume~\bibinfo{volume}{36}, \bibinfo{year}{2022},
  pp. \bibinfo{pages}{780--789}.
%Type = Inproceedings
\bibitem[{Osokin et~al.(2020)Osokin, Sumin, and Lomakin}]{osokin2020os2d}
\bibinfo{author}{A.~Osokin}, \bibinfo{author}{D.~Sumin},
  \bibinfo{author}{V.~Lomakin},
\newblock \bibinfo{title}{Os2d: One-stage one-shot object detection by matching
  anchor features},
\newblock in: \bibinfo{booktitle}{Proceedings of the European conference on
  computer vision}, \bibinfo{organization}{Springer}, \bibinfo{year}{2020}, pp.
  \bibinfo{pages}{635--652}.
%Type = Article
\bibitem[{Hsieh et~al.(2019)Hsieh, Lo, Chen, and Liu}]{hsieh2019one}
\bibinfo{author}{T.-I. Hsieh}, \bibinfo{author}{Y.-C. Lo},
  \bibinfo{author}{H.-T. Chen}, \bibinfo{author}{T.-L. Liu},
\newblock \bibinfo{title}{One-shot object detection with co-attention and
  co-excitation},
\newblock \bibinfo{journal}{Advances in neural information processing systems}
  \bibinfo{volume}{32} (\bibinfo{year}{2019}).
%Type = Inproceedings
\bibitem[{Li et~al.(2021)Li, Yang, Liu, Liu, Ji, and Ye}]{li2021beyond}
\bibinfo{author}{B.~Li}, \bibinfo{author}{B.~Yang}, \bibinfo{author}{C.~Liu},
  \bibinfo{author}{F.~Liu}, \bibinfo{author}{R.~Ji}, \bibinfo{author}{Q.~Ye},
\newblock \bibinfo{title}{Beyond max-margin: Class margin equilibrium for
  few-shot object detection},
\newblock in: \bibinfo{booktitle}{Proceedings of the IEEE conference on
  computer vision and pattern recognition}, \bibinfo{year}{2021}, pp.
  \bibinfo{pages}{7363--7372}.
%Type = Inproceedings
\bibitem[{Chen et~al.(2018)Chen, Wang, Wang, and Qiao}]{chen2018lstd}
\bibinfo{author}{H.~Chen}, \bibinfo{author}{Y.~Wang},
  \bibinfo{author}{G.~Wang}, \bibinfo{author}{Y.~Qiao},
\newblock \bibinfo{title}{Lstd: A low-shot transfer detector for object
  detection},
\newblock in: \bibinfo{booktitle}{Proceedings of the AAAI conference on
  artificial intelligence}, volume~\bibinfo{volume}{32}, \bibinfo{year}{2018},
  pp. \bibinfo{pages}{2836--2843}.
%Type = Inproceedings
\bibitem[{Zhang and Wang(2021)}]{zhang2021hallucination}
\bibinfo{author}{W.~Zhang}, \bibinfo{author}{Y.-X. Wang},
\newblock \bibinfo{title}{Hallucination improves few-shot object detection},
\newblock in: \bibinfo{booktitle}{Proceedings of the IEEE conference on
  computer vision and pattern recognition}, \bibinfo{year}{2021}, pp.
  \bibinfo{pages}{13008--13017}.
%Type = Inproceedings
\bibitem[{Li et~al.(2021)Li, Zhu, Cheng, Wang, Teo, Xiang, Vadakkepat, and
  Lee}]{li2021few}
\bibinfo{author}{Y.~Li}, \bibinfo{author}{H.~Zhu}, \bibinfo{author}{Y.~Cheng},
  \bibinfo{author}{W.~Wang}, \bibinfo{author}{C.~S. Teo},
  \bibinfo{author}{C.~Xiang}, \bibinfo{author}{P.~Vadakkepat},
  \bibinfo{author}{T.~H. Lee},
\newblock \bibinfo{title}{Few-shot object detection via classification
  refinement and distractor retreatment},
\newblock in: \bibinfo{booktitle}{Proceedings of the IEEE conference on
  computer vision and pattern recognition}, \bibinfo{year}{2021}, pp.
  \bibinfo{pages}{15395--15403}.
%Type = Inproceedings
\bibitem[{Wu et~al.(2021)Wu, Han, Zhu, and Yang}]{wu2021universal}
\bibinfo{author}{A.~Wu}, \bibinfo{author}{Y.~Han}, \bibinfo{author}{L.~Zhu},
  \bibinfo{author}{Y.~Yang},
\newblock \bibinfo{title}{Universal-prototype enhancing for few-shot object
  detection},
\newblock in: \bibinfo{booktitle}{Proceedings of the IEEE international
  conference on computer vision}, \bibinfo{year}{2021}, pp.
  \bibinfo{pages}{9567--9576}.
%Type = Inproceedings
\bibitem[{Zhu et~al.(2021)Zhu, Chen, Ahmed, Shen, and
  Savvides}]{zhu2021semantic}
\bibinfo{author}{C.~Zhu}, \bibinfo{author}{F.~Chen},
  \bibinfo{author}{U.~Ahmed}, \bibinfo{author}{Z.~Shen},
  \bibinfo{author}{M.~Savvides},
\newblock \bibinfo{title}{Semantic relation reasoning for shot-stable few-shot
  object detection},
\newblock in: \bibinfo{booktitle}{Proceedings of the IEEE conference on
  computer vision and pattern recognition}, \bibinfo{year}{2021}, pp.
  \bibinfo{pages}{8782--8791}.
%Type = Article
\bibitem[{Grill et~al.(2020)Grill, Strub, Altch{\'e}, Tallec, Richemond,
  Buchatskaya, Doersch, Avila~Pires, Guo, Gheshlaghi~Azar
  et~al.}]{grill2020bootstrap}
\bibinfo{author}{J.-B. Grill}, \bibinfo{author}{F.~Strub},
  \bibinfo{author}{F.~Altch{\'e}}, \bibinfo{author}{C.~Tallec},
  \bibinfo{author}{P.~Richemond}, \bibinfo{author}{E.~Buchatskaya},
  \bibinfo{author}{C.~Doersch}, \bibinfo{author}{B.~Avila~Pires},
  \bibinfo{author}{Z.~Guo}, \bibinfo{author}{M.~Gheshlaghi~Azar}, et~al.,
\newblock \bibinfo{title}{Bootstrap your own latent-a new approach to
  self-supervised learning},
\newblock \bibinfo{journal}{Advances in neural information processing systems}
  \bibinfo{volume}{33} (\bibinfo{year}{2020}) \bibinfo{pages}{21271--21284}.
%Type = Inproceedings
\bibitem[{He et~al.(2020)He, Fan, Wu, Xie, and Girshick}]{he2020momentum}
\bibinfo{author}{K.~He}, \bibinfo{author}{H.~Fan}, \bibinfo{author}{Y.~Wu},
  \bibinfo{author}{S.~Xie}, \bibinfo{author}{R.~Girshick},
\newblock \bibinfo{title}{Momentum contrast for unsupervised visual
  representation learning},
\newblock in: \bibinfo{booktitle}{Proceedings of the IEEE conference on
  computer vision and pattern recognition}, \bibinfo{year}{2020}, pp.
  \bibinfo{pages}{9729--9738}.
%Type = Article
\bibitem[{Goodfellow et~al.(2020)Goodfellow, Pouget-Abadie, Mirza, Xu,
  Warde-Farley, Ozair, Courville, and Bengio}]{goodfellow2020generative}
\bibinfo{author}{I.~Goodfellow}, \bibinfo{author}{J.~Pouget-Abadie},
  \bibinfo{author}{M.~Mirza}, \bibinfo{author}{B.~Xu},
  \bibinfo{author}{D.~Warde-Farley}, \bibinfo{author}{S.~Ozair},
  \bibinfo{author}{A.~Courville}, \bibinfo{author}{Y.~Bengio},
\newblock \bibinfo{title}{Generative adversarial networks},
\newblock \bibinfo{journal}{Communications of the ACM} \bibinfo{volume}{63}
  (\bibinfo{year}{2020}) \bibinfo{pages}{139--144}.
%Type = Inproceedings
\bibitem[{Shrivastava et~al.(2016)Shrivastava, Gupta, and
  Girshick}]{shrivastava2016training}
\bibinfo{author}{A.~Shrivastava}, \bibinfo{author}{A.~Gupta},
  \bibinfo{author}{R.~Girshick},
\newblock \bibinfo{title}{Training region-based object detectors with online
  hard example mining},
\newblock in: \bibinfo{booktitle}{Proceedings of the IEEE conference on
  computer vision and pattern recognition}, \bibinfo{year}{2016}, pp.
  \bibinfo{pages}{761--769}.
%Type = Article
\bibitem[{Wang et~al.(2021)Wang, Chen, and Zhu}]{wang2021survey}
\bibinfo{author}{X.~Wang}, \bibinfo{author}{Y.~Chen}, \bibinfo{author}{W.~Zhu},
\newblock \bibinfo{title}{A survey on curriculum learning},
\newblock \bibinfo{journal}{IEEE Transactions on Pattern Analysis and Machine
  Intelligence} \bibinfo{volume}{44} (\bibinfo{year}{2021})
  \bibinfo{pages}{4555--4576}.
%Type = Article
\bibitem[{Kumar et~al.(2010)Kumar, Packer, and Koller}]{kumar2010self}
\bibinfo{author}{M.~Kumar}, \bibinfo{author}{B.~Packer},
  \bibinfo{author}{D.~Koller},
\newblock \bibinfo{title}{Self-paced learning for latent variable models},
\newblock \bibinfo{journal}{Advances in neural information processing systems}
  \bibinfo{volume}{23} (\bibinfo{year}{2010}).
%Type = Inproceedings
\bibitem[{Jiang et~al.(2015)Jiang, Meng, Zhao, Shan, and
  Hauptmann}]{jiang2015self}
\bibinfo{author}{L.~Jiang}, \bibinfo{author}{D.~Meng},
  \bibinfo{author}{Q.~Zhao}, \bibinfo{author}{S.~Shan},
  \bibinfo{author}{A.~Hauptmann},
\newblock \bibinfo{title}{Self-paced curriculum learning},
\newblock in: \bibinfo{booktitle}{Proceedings of the AAAI conference on
  artificial intelligence}, \bibinfo{year}{2015}, pp.
  \bibinfo{pages}{2694--2700}.
%Type = Article
\bibitem[{Ren et~al.(2015)Ren, He, Girshick, and Sun}]{ren2015faster}
\bibinfo{author}{S.~Ren}, \bibinfo{author}{K.~He},
  \bibinfo{author}{R.~Girshick}, \bibinfo{author}{J.~Sun},
\newblock \bibinfo{title}{Faster r-cnn: Towards real-time object detection with
  region proposal networks},
\newblock \bibinfo{journal}{Advances in neural information processing systems}
  \bibinfo{volume}{28} (\bibinfo{year}{2015}).
%Type = Article
\bibitem[{Yang et~al.(2020)Yang, Wei, Shi, and Li}]{yang2020restoring}
\bibinfo{author}{Y.~Yang}, \bibinfo{author}{F.~Wei}, \bibinfo{author}{M.~Shi},
  \bibinfo{author}{G.~Li},
\newblock \bibinfo{title}{Restoring negative information in few-shot object
  detection},
\newblock \bibinfo{journal}{Advances in neural information processing systems}
  \bibinfo{volume}{33} (\bibinfo{year}{2020}) \bibinfo{pages}{3521--3532}.
%Type = Inproceedings
\bibitem[{Han et~al.(2021)Han, He, Huang, Ma, and Chang}]{han2021query}
\bibinfo{author}{G.~Han}, \bibinfo{author}{Y.~He}, \bibinfo{author}{S.~Huang},
  \bibinfo{author}{J.~Ma}, \bibinfo{author}{S.-F. Chang},
\newblock \bibinfo{title}{Query adaptive few-shot object detection with
  heterogeneous graph convolutional networks},
\newblock in: \bibinfo{booktitle}{Proceedings of the IEEE international
  conference on computer vision}, \bibinfo{year}{2021}, pp.
  \bibinfo{pages}{3263--3272}.
%Type = Inproceedings
\bibitem[{He et~al.(2016)He, Zhang, Ren, and Sun}]{he2016deep}
\bibinfo{author}{K.~He}, \bibinfo{author}{X.~Zhang}, \bibinfo{author}{S.~Ren},
  \bibinfo{author}{J.~Sun},
\newblock \bibinfo{title}{Deep residual learning for image recognition},
\newblock in: \bibinfo{booktitle}{Proceedings of the IEEE conference on
  computer vision and pattern recognition}, \bibinfo{year}{2016}, pp.
  \bibinfo{pages}{770--778}.

\end{thebibliography}
